\documentclass[runningheads]{llncs}
\usepackage{graphicx}
\usepackage{multirow}
\usepackage{subfig}
\usepackage{amsmath,amsfonts,amssymb}
\usepackage[sorting=none, style=numeric]{biblatex}
\usepackage{enumerate}
\usepackage[linesnumbered,ruled,vlined]{algorithm2e}

\usepackage{array}
\newcolumntype{P}[1]{>{\centering\arraybackslash}p{#1}}

\usepackage{microtype}
\usepackage{colortbl}
\usepackage[hidelinks]{hyperref}

\usepackage{xcolor}
\usepackage{multicol}
\usepackage{pifont}
\usepackage{newunicodechar}
\newunicodechar{✓}{\ding{51}}
\newunicodechar{✗}{\ding{55}}
\usepackage{tabularray}

\begin{document}

\title{%
    \makebox[\textwidth][c]{Predicting Cardiopulmonary Exercise Testing Outcomes} \\
    \makebox[\textwidth][c]{in Congenital Heart Disease Through Multi-modal} \\
    \makebox[\textwidth][c]{Data Integration and Geometric Learning} \\
}

\titlerunning{Digitisation and Linkage of PDF formatted 12-lead ECGs in ACHD}

\author{Muhammet Alkan\inst{1} 
\and Gruschen Veldtman\inst{2,3}
\and Fani Deligianni\inst{1}\thanks{Corresponding author email: fani.deligianni@glasgow.ac.uk}}

\authorrunning{F. Deligianni et al.}

\institute{School of Computing Science at University of Glasgow, Glasgow, Scotland, UK \and
Golden Jubilee National Hospital, Glasgow, Scotland, UK \and Helen DeVos Children's Hospital, Corewell Health, Hospital, Michigan, USA
}

\maketitle

\section*{Abstract}

Cardiopulmonary exercise testing (CPET) provides a comprehensive assessment of functional capacity by measuring key physiological variables including oxygen consumption ($VO_2$), carbon dioxide production ($VCO_2$), and pulmonary ventilation ($VE$) during exercise. Previous research has established that parameters such as peak $VO_2$ and $VE/VCO_2$ ratio serve as robust predictors of mortality risk in chronic heart failure patients. In this study, we leverage CPET variables as surrogate mortality endpoints for patients with Congenital Heart Disease (CHD). To our knowledge, this represents the first successful implementation of an advanced machine learning approach that predicts CPET outcomes by integrating electrocardiograms (ECGs) with information derived from clinical letters.
Our methodology began with extracting unstructured patient information—including intervention history, diagnoses, and medication regimens—from clinical letters using natural language processing techniques, organizing this data into a structured database. We then digitized ECGs to obtain quantifiable waveforms and established comprehensive data linkages. The core innovation of our approach lies in exploiting the Riemannian geometric properties of covariance matrices derived from both 12-lead ECGs and clinical text data to develop robust regression and classification models. Through extensive ablation studies, we demonstrated that the integration of ECG signals with clinical documentation, enhanced by covariance augmentation techniques in Riemannian space, consistently produced superior predictive performance compared to conventional approaches.

\newpage
\section{Introduction}

The 12-lead electrocardiogram (ECG) analysis remains a cornerstone in cardiac diagnostics and prognosis, offering unique advantages through its accessibility and high spatio-temporal resolution of cardiac function.  Its particular value in congenital heart disease (CHD) lies in its ability to reflect underlying anatomical abnormalities through distinct ECG patterns \cite{waldmann2020understanding}. Several deep learning techniques have been proposed to classify cardiac rhythms and estimate the risk of adverse effects \cite{ribeiro2020automatic}. These methods showed impressive results with large datasets that include millions of patients and ECG recordings. However, it is not clear how they can extend in relatively rare and extremely heterogeneous cases.\\

Congenital heart disease presents a unique challenge in this context. As a condition present from birth, CHD affects approximately 1\% of newborns globally, translating to roughly 1.2 million cases annually worldwide \cite{british2024global}. In the United Kingdom alone, the incidence rate of about 1 in every 100 births results in approximately 4,600 new cases each year \cite{petersen2003congenital}. The prevalence of CHD emphasises the importance of early detection and treatment in order to improve the outcomes of those affected. CHD encompasses a diverse range of structural and functional cardiac abnormalities present at birth. Symptoms can range from subtle symptoms like rapid heartbeat and breathing difficulties to more severe feeding problems in infants. While the aetiology often remains unclear, known risk factors include genetic conditions, maternal infections, certain medications, and poorly controlled diabetes during pregnancy.\\

The condition's heterogeneity poses particular challenges for machine learning approaches, as patients present with genetic defects that differ substantially from cardiac abnormalities developing later in life. This fundamental difference limits the effectiveness of deep learning methods developed for broader populations, primarily due to two factors: the scarcity of large-scale representative data and the extreme physiological variations in both anatomy and function. In this context, cardiopulmonary exercise testing (CPET) emerges as a valuable tool for patient's assessment and management \cite{cifra2024cardiopulmonary, constantine2022cardiopulmonary}. As a specialized exercise protocol, CPET provides comprehensive insights into a patient's functional capacity and exercise ability. Its clinical significance lies in its ability to objectively measure exercise capacity, helping identify at-risk patients.  The test yields various physiological parameters, including oxygen consumption, that can be compared against reference values from healthy populations. The widespread adoption of CPET is evident in its implementation across 68\% of UK hospital departments \cite{reeves2018perioperative}, where it aids in evaluating patients before major procedures or surgeries.\\

In this study, we present several novel contributions to advance the field of CHD patient monitoring and risk assessment. First, we develop an innovative machine learning approach that fuses information from both ECG signals and clinical letters, creating a more comprehensive patient profile. Second, we introduce a sophisticated covariance mixing regularization technique that leverages Riemannian geometry to handle the inherent challenges of small, imbalanced datasets common in CHD cases. Third, we demonstrate the effectiveness of using CPET variables ($VO_2$ and $VE/VCO_2$) as surrogate outcomes for mortality prediction, providing a more nuanced approach to risk assessment than traditional binary outcomes. Finally, we validate our methodology through extensive ablation studies, showing that the integration of multiple data sources and our novel augmentation technique significantly improves predictive performance compared to conventional approaches.

\section{Background}

\subsection{Cardiopulmonary exercise testing for risk prediction in cardiovascular diseases}

CPET provides invaluable insights into a patient's cardiorespiratory fitness by measuring key variables like oxygen consumption, carbon dioxide production, heart rate and ventilatory parameters \cite{wadey2022role}. Oxygen consumption ($VO_2$) is a crucial cardiopulmonary exercise variable  that serves as a reliable predictor of mortality and morbidity in patients with CHD. This concept is rooted in the work of Hill et al., who introduced the idea that there is an individual exercise intensity at which $VO_2$ no longer increases \cite{hill1923muscular}. Consequently, $VO_2$ peak represents the limit of cardiorespiratory capacity and is a strong indicator of whether individuals reach maximal conditions at the end of a CPET. The normal range for $VO_2$ peak is typically between 25-35 ml/kg/min. Studies have shown that lower peak oxygen consumption ($VO_2$ peak) values during CPET are associated with higher risks of mortality, especially in patients undergoing major surgeries \cite{reeves2018cardiopulmonary}.\\

On the other hand, $VO_2$ \%pred is the percentage of the predicted maximum oxygen consumption based on factors such as age, sex, and height. This measure is useful for assessing ventilatory efficiency, and may indicate potential respiratory and/or cardiac limitations. The normal range for $VO_2$ \%pred is typically within the range of 60 to 85 percent. Another important variable is the $VE/VCO_2$ ratio, which measures the relationship between pulmonary ventilation (VE) and carbon dioxide production ($VCO_2$). $VCO_2$ represents the maximum amount of carbon dioxide that can be produced during the exercise. Thus, $VE/VCO_2$ ratio can be summarised as the required ventilation to eliminate the $CO_2$ produced during the test. The normal range for $VE/VCO_2$ is typically between 20 and 30. Previous works have demonstrated that these variables are independent predictors of a high mortality risk in patients with Chronic Heart Failure (CHF) \cite{nanas2006ve, abella2020cardiopulmonary}. Nanas et al. confirmed that the $VE/VCO_2$ slope is a strong, independent predictor of high mortality risk in CHF patients \cite{nanas2006ve}. Another study demonstrated that the $VE/VCO_2$ slope is a significant predictor of cardiac-related hospitalizations in CHF patients \cite{shen2015ve}. $VO_2$ and $VE/VCO_2$ slope values have been correlated with long-term mortality risk in adults with CHD, with increased risk observed in cases of low $VO_2$, low heart rate reserve, and high $VE/VCO_2$ in non-cyanotic heart diseases \cite{abella2020cardiopulmonary}.

Given the challenges of directly modeling mortality risk due to low prevalence, CPET offers more than just a pragmatic statistical solution. It provides a dynamic approach to patient monitoring that extends beyond mortality, which is an extreme outcome. By tracking continuous variables such as $VO_2$ peak and $VE/VCO_2$ ratio, clinicians can capture subtle changes in a patient's physiological functioning, allowing for early detection of declining health, personalized intervention strategies, and more proactive medical management. This approach enables healthcare providers to assess and predict cardiovascular risk with greater sensitivity, tracking the patient's functional capacity and potential health trajectories long before critical events might occur. Recent advancements also suggest the potential of predicting CPET outcomes from ECG data using machine learning algorithms, which could further streamline the assessment process \cite{otto2020cardiopulmonary}.

\subsection{Machine Learning for ECG-based CHD Classification and Risk Prediction}

Electrocardiogram (ECG) is a significant tool in diagnosing and managing congenital heart defects (CHD). It provides essential diagnostic and prognostic information, revealing heart blocks and defects that may be missed clinically \cite{eisenberg1941congenital,waldmann2020understanding}. The severity of CHD often correlates with abnormal ECG patterns, making it a valuable tool for detecting conditions like atrial septal defect (ASD) and tetralogy of Fallot (ToF), which represent significant portions of adult CHD cases (30\% ASD, 10\% ToF).\\

Classification studies in CHD often focus on identifying and categorizing various heart conditions using ECG data \cite{liu2021deep}. Recent studies employ deep learning (DL) techniques to detect heartbeats, annotate ECG signals, and classify arrhythmias. For instance, Vullings et al. used a deep neural network to classify fetal vectorcardiograms (VCG) as healthy or CHD \cite{vullings2019fetal}. Similarly, Du et al. applied a Residual Network to classify child ECGs, using a dataset of 68,969 ECGs \cite{du2020recognition}. Liang et al. utilized a Residual Network (RN) with cardiac cycle segmentation to improve the detection of CHD with 72,626 child ECGs \cite{du2021recognition}. Kim et al. used a Long Short-Term Memory (LSTM) network to classify heart disease with ECGs \cite{liu2018classification}. Yuan et al. compared different models, including Convolutional Neural Networks (CNNs), Recurrent Neural Networks (RNNs), and Multilayer Neural Networks (MNNs), for ECG classification in CHD patients, finding that MNNs performed best \cite{yuan2021classification}. These studies highlight the potential of machine learning in improving CHD detection rates.\\

Clinical models for predicting mortality and other outcomes in CHD patients have been explored using various approaches. For example, Diller et al. developed a deep learning architecture incorporating ECG parameters along with laboratory and exercise data to categorize diagnostic groups and disease complexity in adult CHD patients \cite{diller2019machine}. Rather than using the raw ECG signals, they only included ECG parameters such as resting heart rate (b.p.m.), QRS duration (ms), and QTc duration (ms) along with laboratory and exercise parameters. By utilizing 13,649 ECGs and 44,421 medical reports of 10,019 adult patients (age 36.3 ± 17.3 years), their model achieved an accuracy of 0.91, 0.97, and 0.90 on categorized diagnosis, disease complexity, and NYHA class. A recent study \cite{mayourian2024electrocardiogram} by researchers at Boston Children's Hospital has further advanced the potential of AI-enhanced ECG analysis for mortality prediction in CHD patients. Their innovative approach addresses a critical gap in risk stratification across patient lifespans. By developing a convolutional neural network trained on 112,804 ECGs from patients aged 0-85 years, the researchers achieved a remarkable area under the receiver operating characteristic curve of 0.79. Notably, the model outperformed traditional clinical markers like age, QRS duration, and left ventricular ejection fraction. In summary, research on congenital heart disease (CHD) using ECG data has been limited to binary classification methods, whether studying pediatric or adult populations. Most investigations have focused exclusively on ECG measurements, with just one study integrating multiple data types. Furthermore, this single multi-modal study utilised derived ECG parameters rather than analyzing raw ECG signals directly.

\section{Methods}

This study aims to utilise $VO_2$ and $VE/VCO_2$ during CPET as a means of predicting mortality in patients with CHD. These surrogate outcomes have demonstrated their efficacy as independent indicators of a high mortality risk in given demographic. A covariance mixing regularisation technique for augmentation was employed for both data types, ECGs and clinical letters. Similar to mixup approach \cite{zhang2017mixup}, this approach performs the interpolation on a Riemannian manifold with respect to the underlying covariance matrices. Building upon our previous work \cite{alkan2024riemannian}, efficiency of the augmentation technique was validated with ablation studies on a dataset of patients with CHD that was extremely small and imbalanced. This technique led to enhancements in both classification and regression problems, which were the best results.

\subsection{Data}

The study encompasses 4,153 12-lead ECGs from 436 patients (194 female) with CHD, who were under regular follow-up at the Scottish Adult Congenital Cardiac Service located at the Golden Jubilee National Hospital in Scotland. The ECG dataset had a mean patient age of 33 years (standard deviation 11.7, interquartile range 23-40). The patient cohort was categorised by their primary cardiac conditions: tetralogy of fallot (ToF, 39.9\%), atrial septal defect (ASD, 17.6\%), and pulmonary atresia (PA, 16.7\%). The ECG recordings were sampled at 500 Hz, with unaligned segments spanning 2.5 seconds and R-peak aligned segments spanning approximately 1 second. And, 12 different leads of the ECGs populated in the following order: DI,DII,DIII,AVR,AVL,AVF,V1,V2,V3,V4,V5,V6.\\

Strict inclusion and exclusion criteria were applied to ensure data integrity. Patients were excluded if they had ECGs showing atrial flutter, atrial fibrillation, or atrioventricular paced rhythms, as the primary aim was to analyse ECGs in sinus rhythm. For patients with multiple anatomic diagnoses, the dominant diagnosis was considered the primary diagnosis. The patient breakdown was as follows: 173 patients with ToF, 77 patients with ASD, 73 patients with PA, 66 patients with Fontan, and 47 patients with Mustard.\\

In addition to ECG data, 595 cardiopulmonary exercise test (CPET) documents from the same patient set were digitised using optical character recognition (OCR) techniques, with all data linked through unique patient identifier numbers. Typically, each patient may have 1 to 3 CPET documents corresponding to different testing sessions conducted over time, depending on the patient's follow-up schedule and clinical needs. For the purpose of this study, we used the CPET document that was closest in date to the ECG data for each patient. This approach ensures that the data reflects the most relevant physiological status corresponding to the ECG measurements, providing a more accurate and aligned analysis. Utilizing the CPET document closest to the ECG date, we aim to minimize potential discrepancies that could arise from changes in the patient's condition over time. Key variables such as $VO_2$ and $VE/VCO_2$ were selected as surrogate outcomes for the assessment of mortality risk. Furthermore, a similar selection approach was also employed for the clinical letters to gather the closest letter in date to the ECG data for each patient, to ensure that the information was as relevant as possible. All the unstructured patient-related information was also extracted from clinical letters for the same set of patients \cite{verma2022development}. This information was obtained using natural language processing and text pre-processing techniques to gain insight into patient history, encompassing details like diagnoses, interventions, and medications. In total, 17 clinical variables were extracted from the letters into columns of a structured table: CHI Number, Name, DOB, Diagnosis, Diagnosis List, Intervention, Intervention List, Medication, Medication List, ESC Classification, Arrhythmia, Clinic Date, Health Board, Postcode, Gender, Height, Weight. With the exclusion of demographic data, list of extracted diagnoses (Diagnosis List), interventions (Intervention List) and medications (Medication List) were combined to gather important information about history of the patients.

\begin{figure}[!h]
    \centering
    \includegraphics[width=0.7\linewidth]{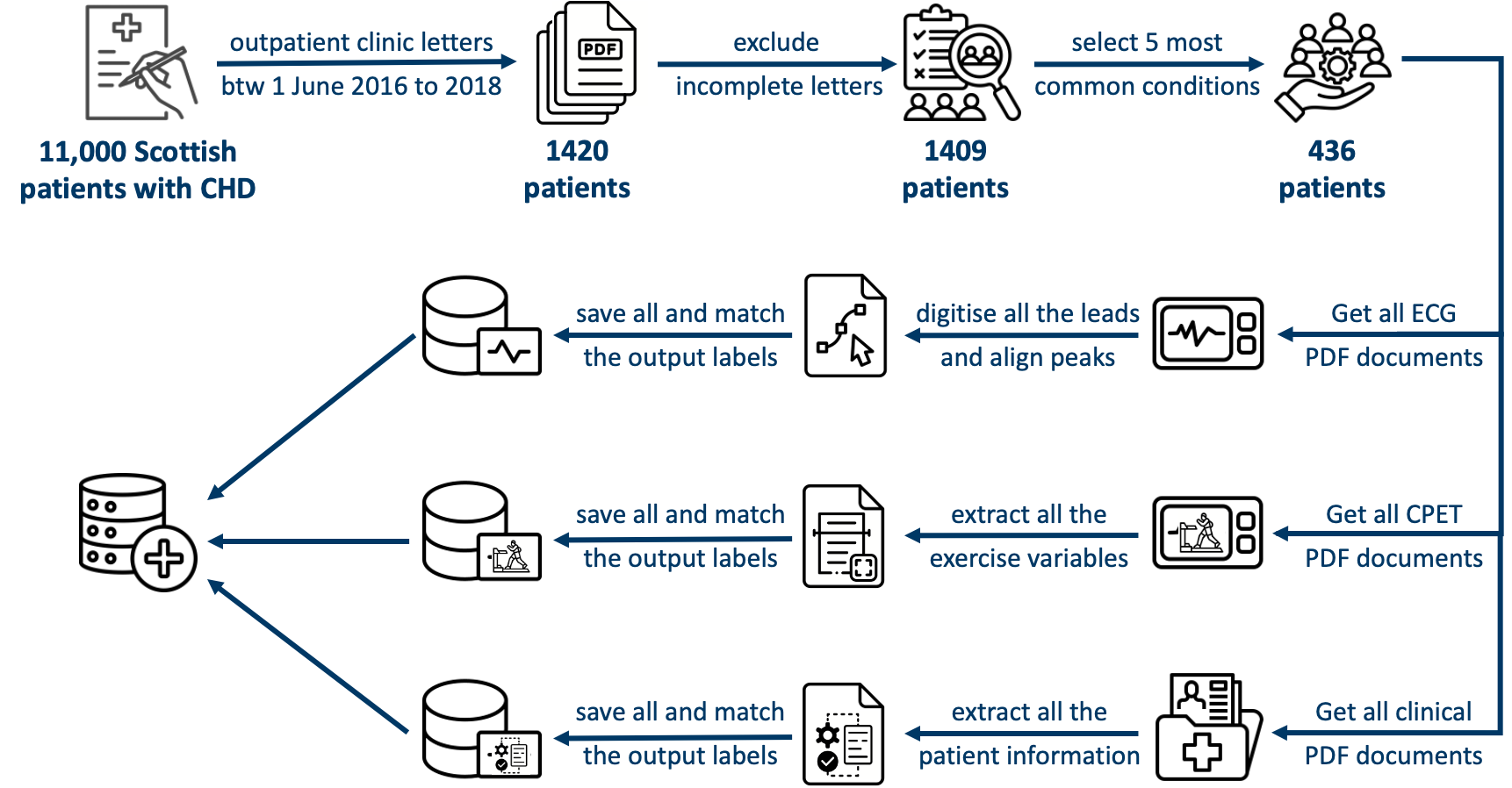}
    \caption{Summary of the data extraction steps}
    \label{fig:enter-label}
\end{figure}

\subsection{Preprocessing}

\begin{figure}[!h]
  \centering
  \begin{tabular}[b]{c}
    \includegraphics[width=.17\linewidth]{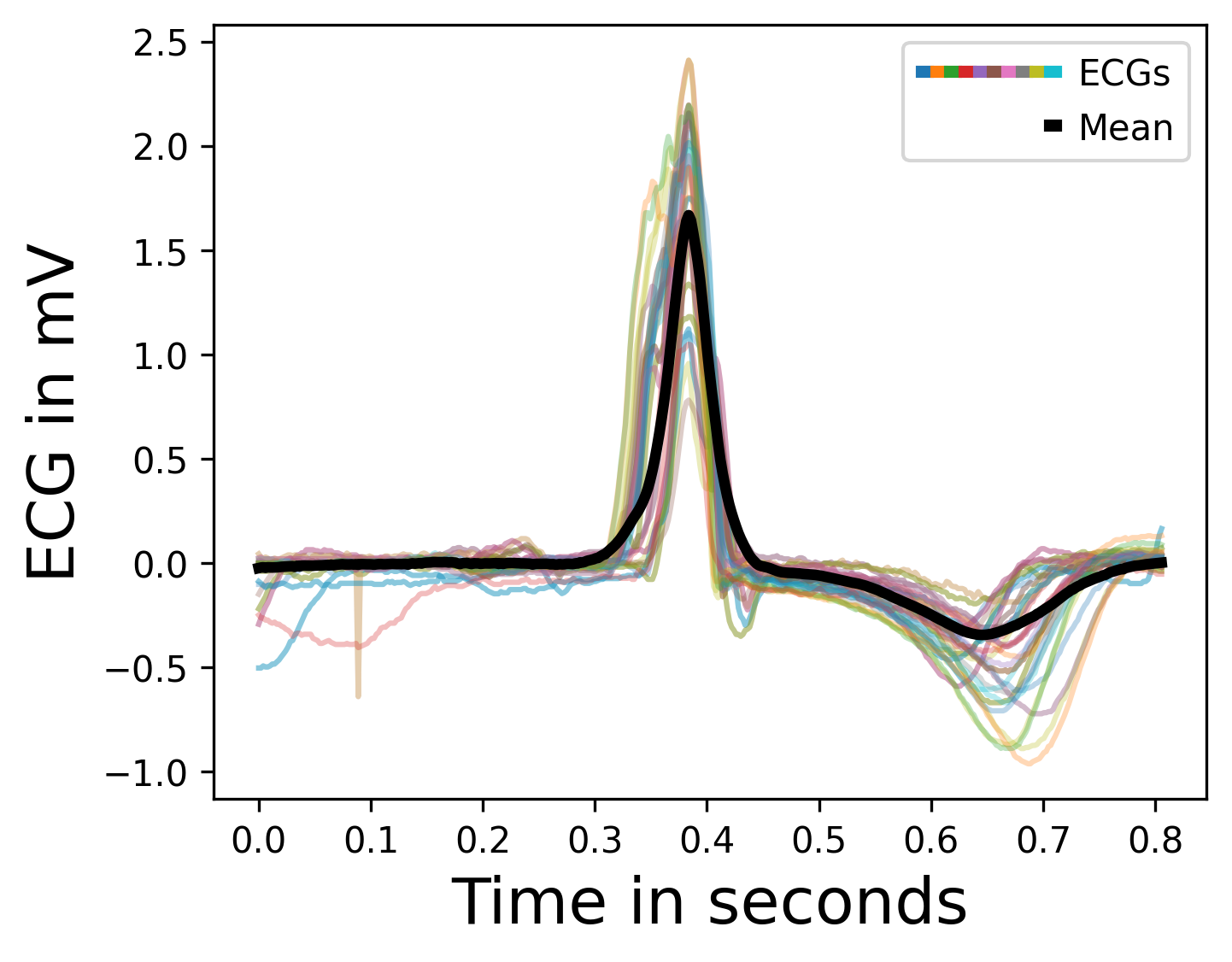} \\
    \scriptsize (a) Mustard,\\ \scriptsize Lead v1
  \end{tabular}
  \begin{tabular}[b]{c}
    \includegraphics[width=.17\linewidth]{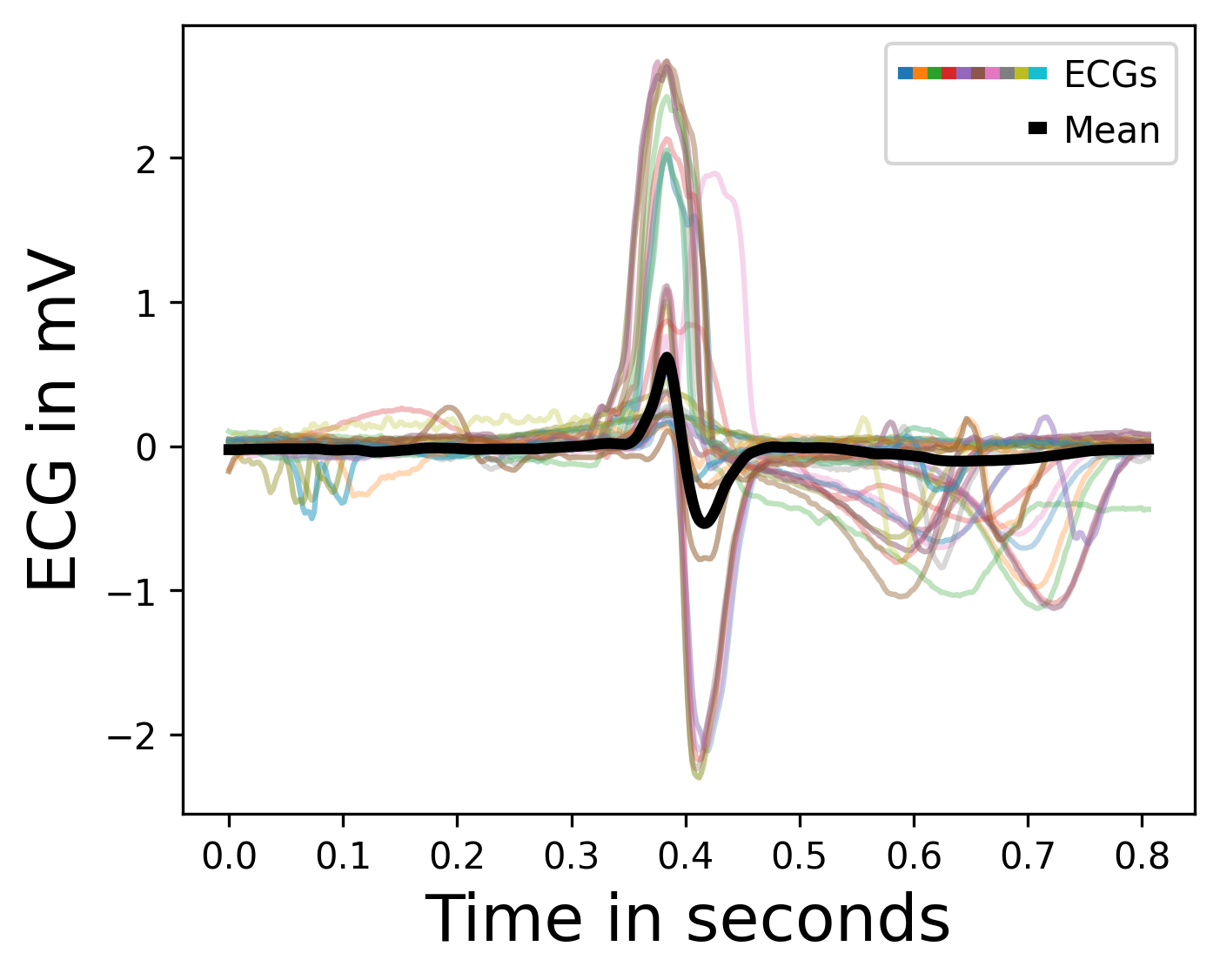} \\
    \scriptsize (b) Fontan,\\ \scriptsize Lead v1
  \end{tabular}
  \begin{tabular}[b]{c}
    \includegraphics[width=.17\linewidth]{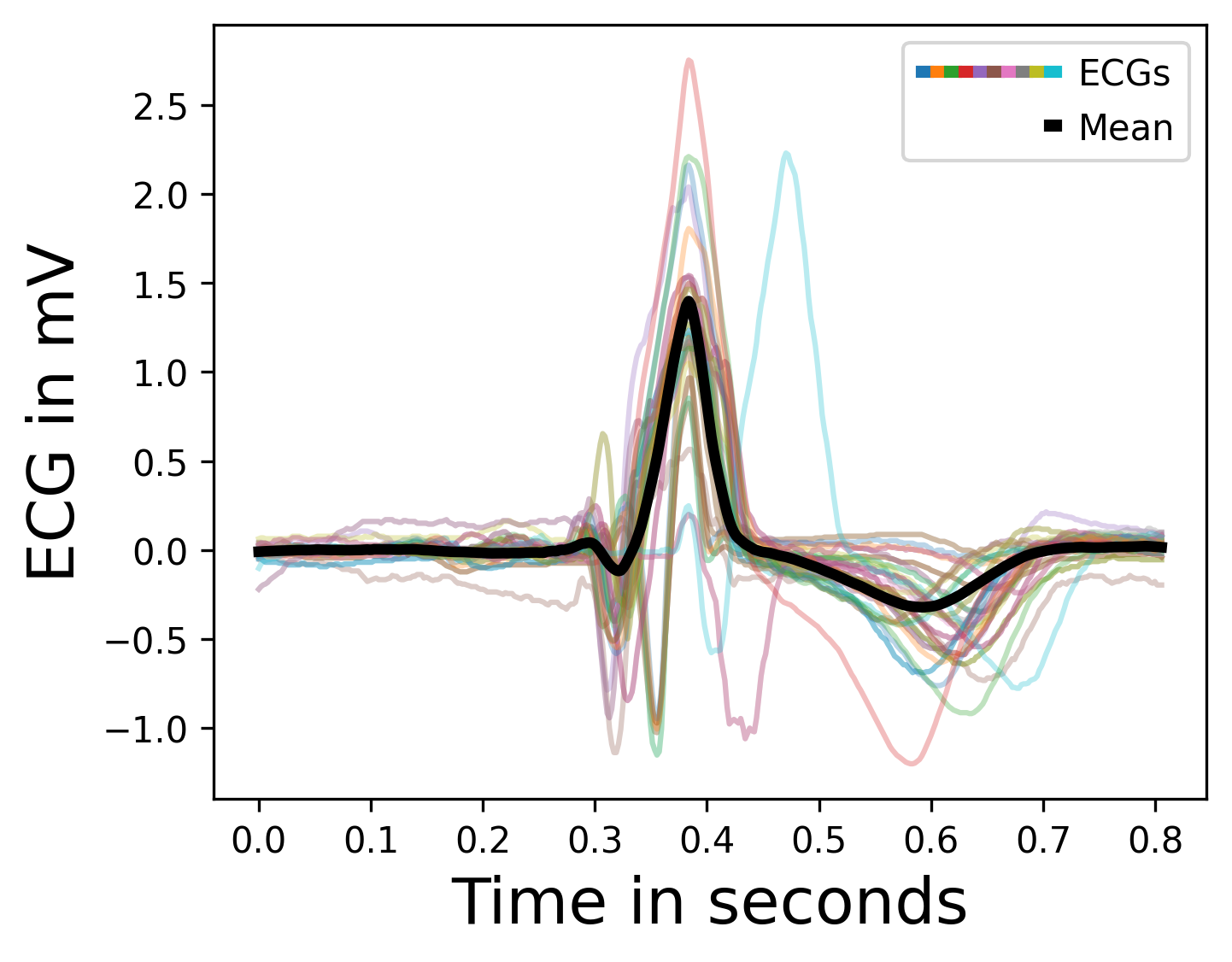} \\
    \scriptsize (b) ToF,\\ \scriptsize Lead v1
  \end{tabular}
  \begin{tabular}[b]{c}
    \includegraphics[width=.17\linewidth]{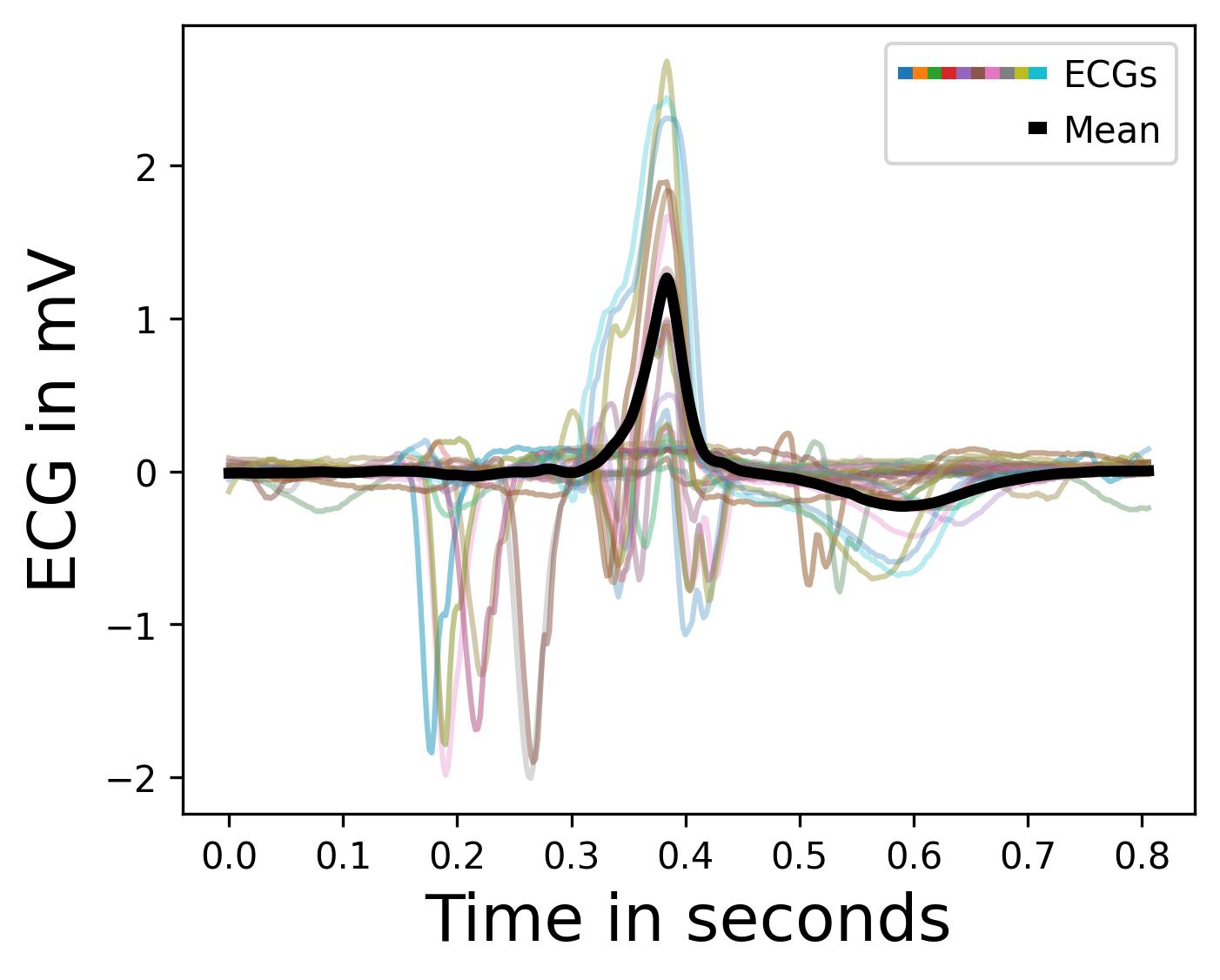} \\
    \scriptsize (b) PA,\\ \scriptsize Lead v1
  \end{tabular}
  \begin{tabular}[b]{c}
    \includegraphics[width=.17\linewidth]{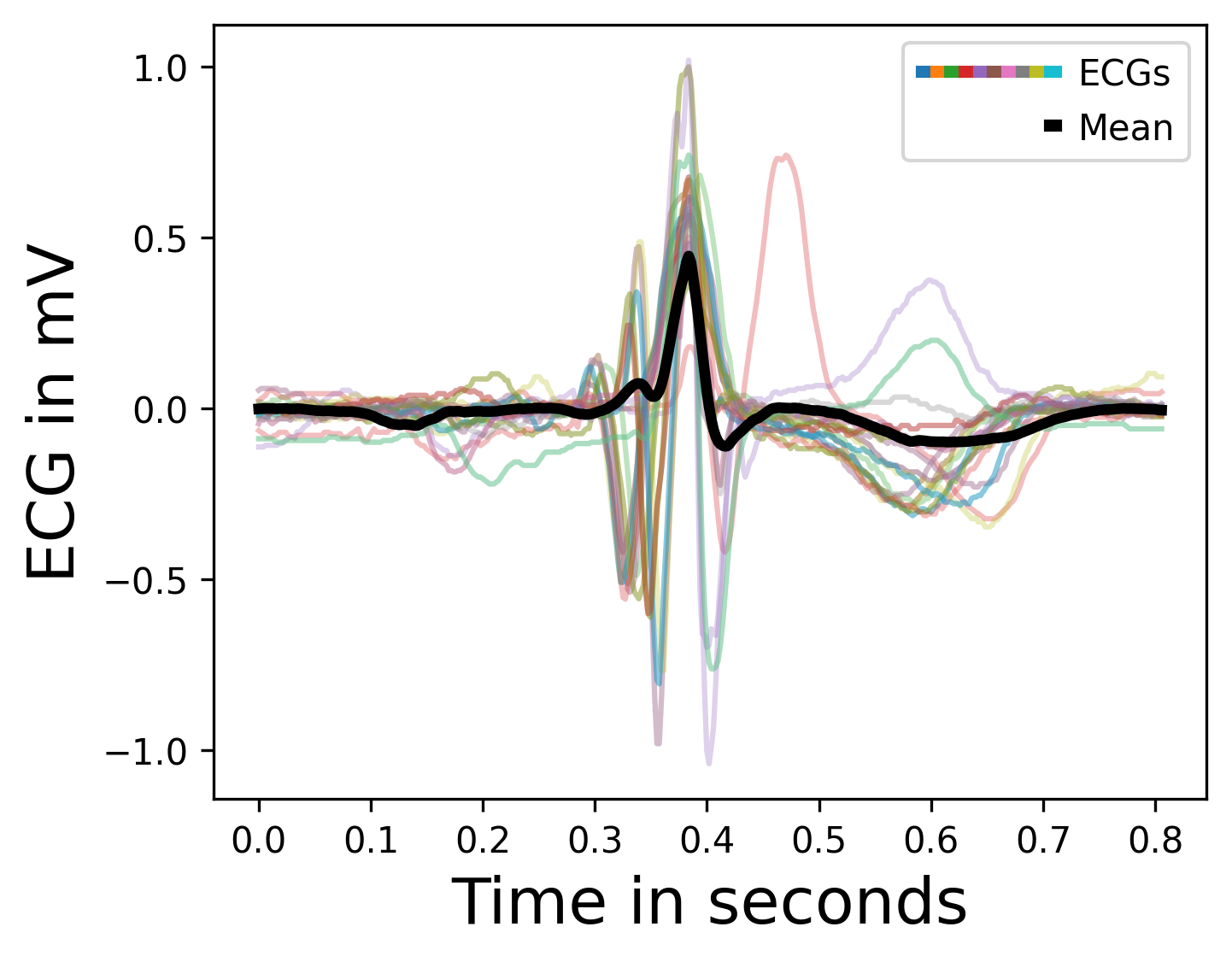} \\
    \scriptsize (b) ASD,\\ \scriptsize Lead v1
  \end{tabular}
  \caption{\centering Aligned ECGs on lead v1 of patients with the specified conditions, with the mean signal outlined in bold}
  \label{fig:beats}
\end{figure}

12-lead ECG data were extracted from ECG PDF documents obtained via the Marquette™ 12SL by GE Healthcare analysis program. We developed an algorithm that automatically digitizes ECG data from PDF documents by leveraging vector drawings, preserving the complete signal information without requiring manual user input \cite{alkan2024digitisation}. To prepare the ECG data for machine learning analysis, we standardized the signals by aligning the R peaks across all heartbeats and leads. This preprocessing step ensured digital synchronization of QRS complexes, creating a uniform temporal reference point across all patient recordings. An example of the average signal of the aligned ECGs for Mustard is shown in Figure \ref{fig:beats} and its characteristic of the underline abnormality. The average length of non-aligned ECGs is 2.5 seconds and R peak aligned ECGs is around 1 second, and they are sampled at a rate of 500 samples per second. The ECG recordings were sampled at 500 Hz, with unaligned segments spanning 2.5 seconds and R-peak aligned segments spanning approximately 1 second.

\vspace{-7mm}

\subsection{Covariance-based ECG analysis}

Our work suggests that in 12-lead ECG signal analysis, covariance matrices can serve as critical mathematical representations that capture the intrinsic statistical relationships and variability within cardiac electrical signals across different leads. These matrices encode crucial information about the signal's spatial and temporal correlations, reflecting the complex interplay between electrical activities of different heart regions \cite{singh2023ecg}. By quantifying the shared variations and dependencies between ECG leads, covariance matrices provide a comprehensive statistical fingerprint of cardiac electrical propagation.\\

Covariance matrices can effectively summarize the multidimensional nature of ECG signals, capturing both amplitude variations and intricate relationships between different cardiac leads. This holistic representation becomes especially significant when attempting to distinguish between different cardiac conditions, as the underlying statistical structure can contain diagnostic signatures that individual lead measurements might obscure.\\

Building upon our previous work \cite{alkan2024riemannian}, we employed a sophisticated Riemannian geometric approach to signal processing. Covariance matrices of ECG signals were first calculated and then mapped to the tangent space using Riemannian distance, a method necessitated by the unique properties of Symmetric Positive Definite (SPD) matrices \cite{barachant2011multiclass}.\\

Unlike traditional Euclidean space analysis, SPD matrices require specialized geometric treatment. Riemannian geometry provides a more nuanced approach to handling these matrices, allowing for precise projection while preserving their inherent mathematical structure \cite{barachant2011multiclass}. The projection onto a common tangent space is governed by the following mathematical transformation \cite{yger2016riemannian}:

\begin{equation}
\label{eq:tangentSpace}
\mathbf{V_i^{C}} = {\rm upper} \left({\bf C}^{-{\frac{1}{2}} } {\rm Log}_{{\bf C}} \left({\bf C}_i \right) {\bf C}^{-{\frac{1}{2}}} \right)
\end{equation}

In this equation, $\bf C_i$ represents the individual covariance matrix being projected onto the tangent space at point $\bf C$, which serves as the Riemannian mean of all covariance matrices. This projection methodology offers critical advantages, enabling more accurate distance metrics between sample covariance matrices and demonstrating exceptional performance in processing high-dimensional neurophysiological data.\\

Following the tangent space projection, each covariance matrix is transformed into a vector $\mathbf{V}$ with dimensions $n\times(n+1)/2$, where $n$ represents the original matrix dimension. Specifically, the projection retains only the upper triangular portion of the resulting symmetric matrix, as defined in Equation \ref{eq:tangentSpace}.\\

The method provides a sophisticated approach to preserving the geometric information inherent in SPD matrices, offering a more refined technique for analyzing complex signal data compared to traditional Euclidean methods.

\subsection{Covariance-based ECG Augmentation}

Advancing beyond traditional data augmentation techniques, we adopt a sophisticated covariance mixing regularisation method that leverages Riemannian geometric principles. Inspired by the linear interpolation approach of mixup \cite{zhang2017mixup}, this technique performs sample interpolation within the Riemannian manifold, specifically tailored to the intrinsic geometry of covariance matrices.\\

The augmentation process begins by sampling an interpolation factor $\alpha$ from a beta distribution constrained to the interval [0, 1]. This sampling strategy allows probabilistic mixing of covariance matrices. Unlike conventional methods that operate in Euclidean space, our approach computes the weighted Riemannian mean by calculating the distance between randomly selected covariance matrices using a Riemannian distance metric.\\

The weighted Riemannian mean is computed by minimizing the sum of squared Riemannian distances to the given Symmetric Positive Definite (SPD) matrices, as formalized in Equation \ref{eq:meanRiemannian}. Here, $w_i$ represents a weight matrix generated using the $\alpha$ value, and $d_R$ represents the Riemannian distances to the SPD matrices.

\begin{equation}
\label{eq:meanRiemannian}
\mathbf{C_{aug}} = \arg \min_{\mathbf{C}} \sum_i w_i \ d_R (\mathbf{C}, \mathbf{C}_i)^2
\end{equation}

\subsection{Fusion of information derived from ECGs and Clinical Letters}

From the initially extracted 17 clinical variables, we focused on three key categories of patient history: diagnoses, interventions, and medications, excluding demographic data. To integrate this clinical history information with the ECG signals in tangent space, we developed a comprehensive text preprocessing pipeline. The pipeline transformed clinical letters into structured numerical representations through the following steps. First, we converted the textual content into a sparse matrix representation, where each row corresponds to a unique clinical letter and each column represents a distinct term found across all letters. This matrix captured term frequencies, quantifying the occurrence patterns of medical terminology within each document. We then concatenated these word-frequency matrices with the tangent space mapping matrices derived from ECG signals to create unified input features for our model. This fusion approach enabled us to simultaneously leverage both the semantic patterns present in clinical documentation and the geometric properties inherent in ECG signals, providing a more comprehensive representation of each patient's cardiac condition. The combined feature space preserved both key features of clinical assessments and the characteristics of electrical cardiac activity.

\begin{figure}[h]
    \centering
    \includegraphics[width=0.85\linewidth]{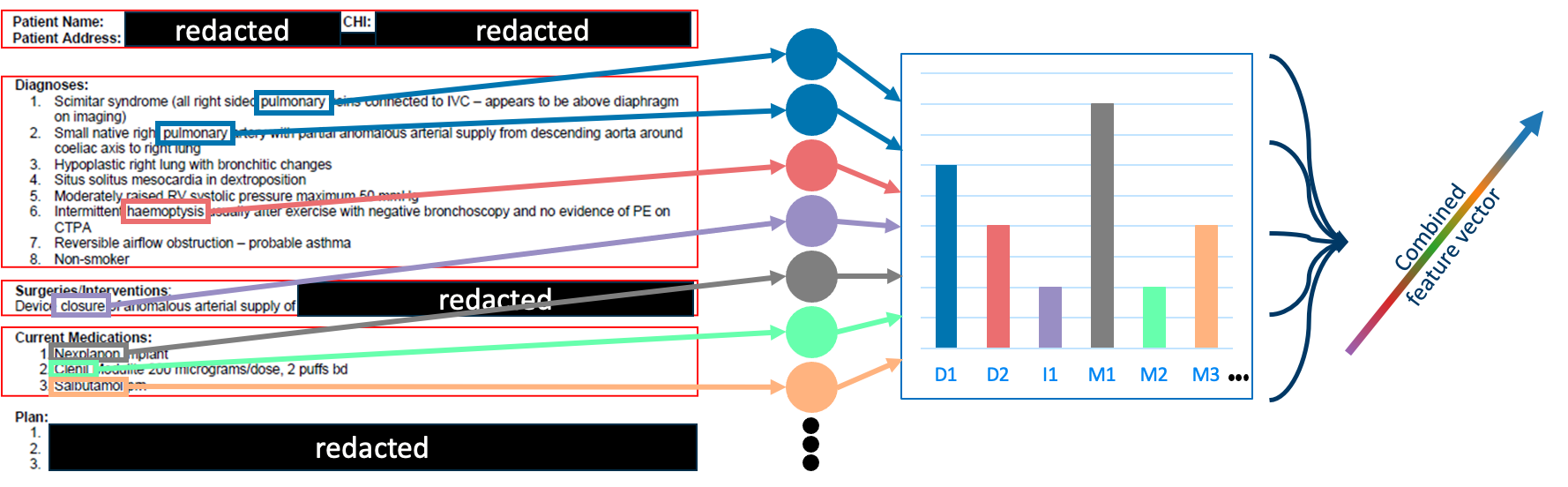}
    \caption{Text to matrix transformation using CountVectorizer}
    \label{fig:countVectorizer}
\end{figure}

\begin{figure}[h]
  \centering
  \begin{tabular}[b]{c}
    \includegraphics[width=.22\linewidth]{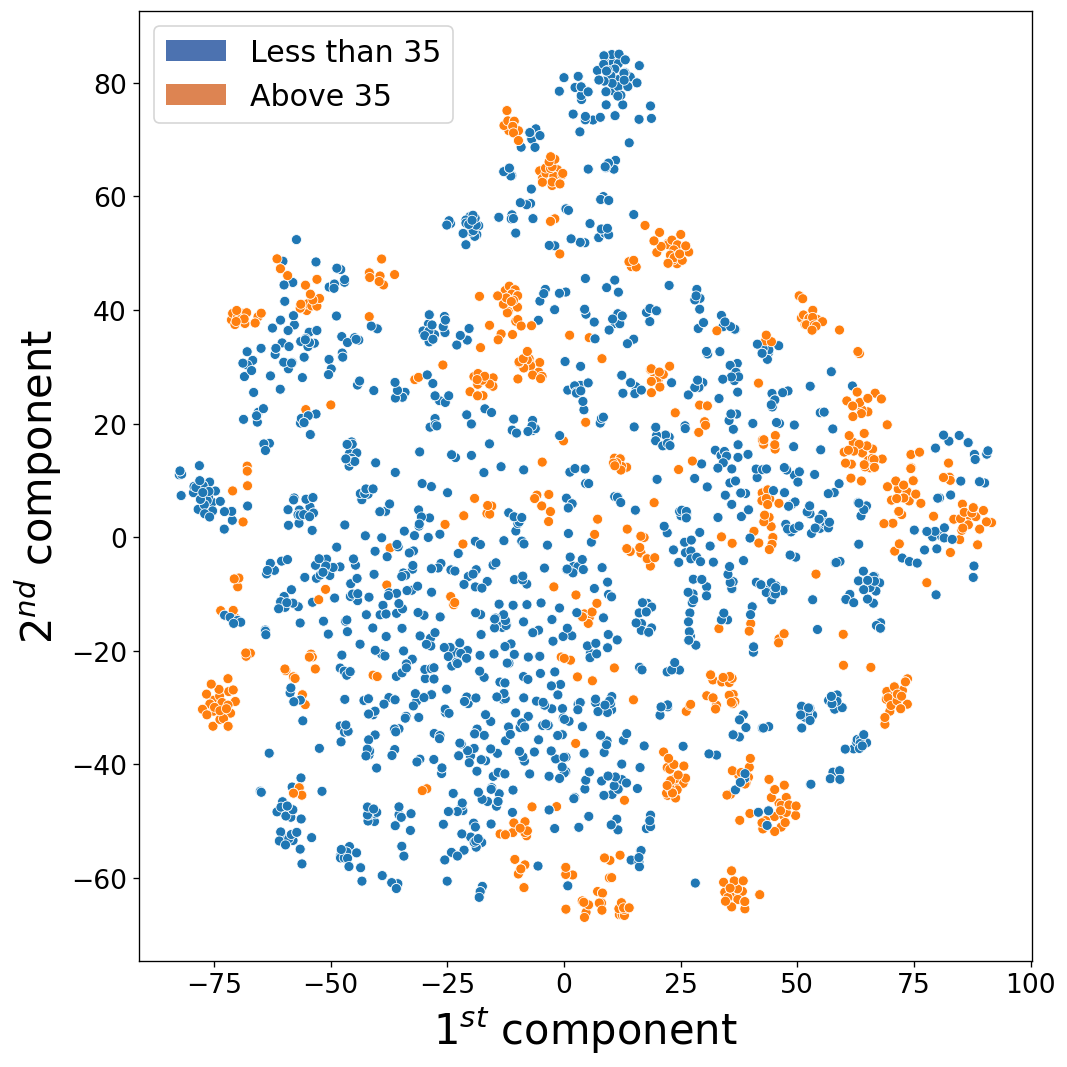} \\
    \scriptsize (a) ECG without\\ \scriptsize augmentations
  \end{tabular}
  \begin{tabular}[b]{c}
    \includegraphics[width=.22\linewidth]{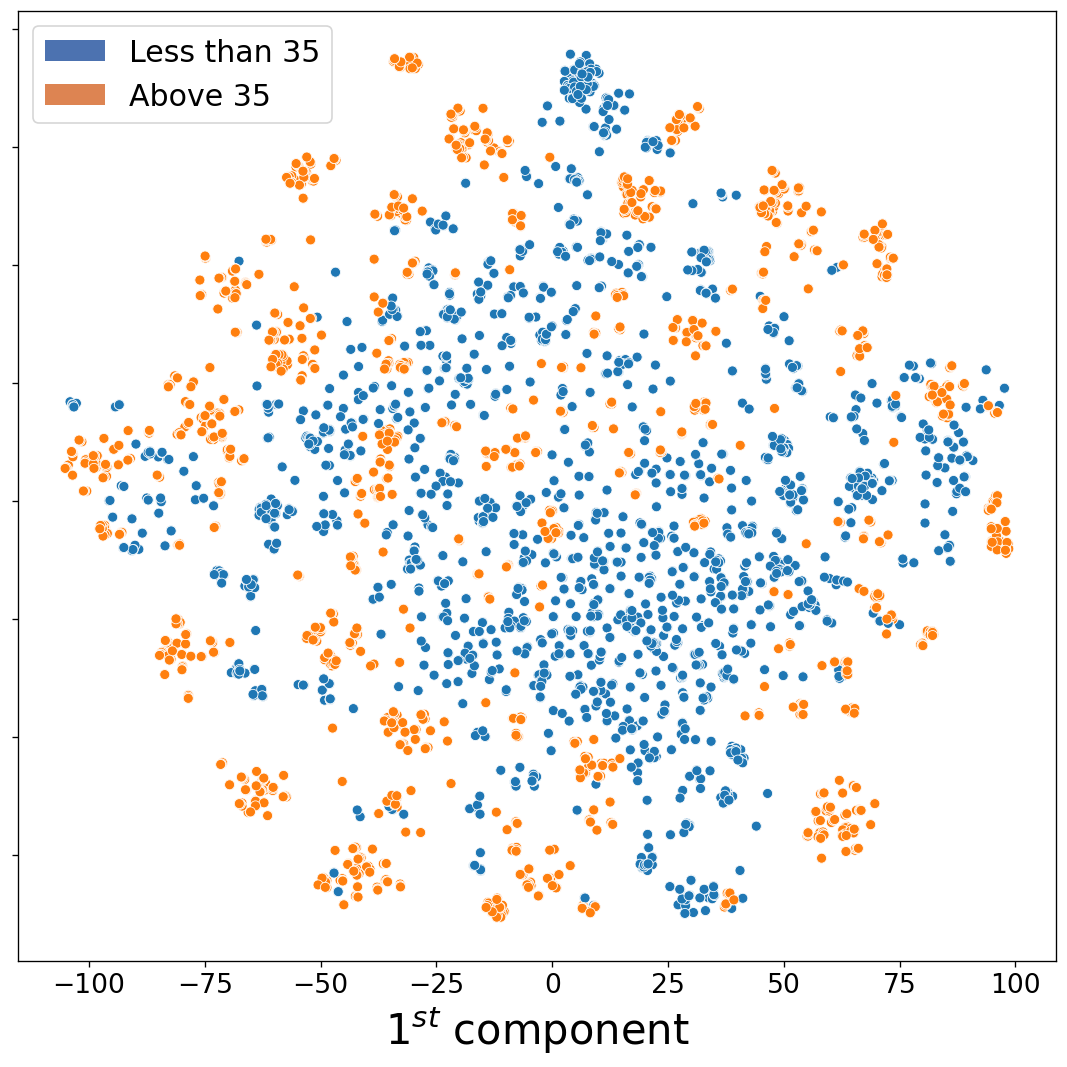} \\
    \scriptsize (b) ECG with\\ \scriptsize augmentations
  \end{tabular}
   \begin{tabular}[b]{c}
    \includegraphics[width=.22\linewidth]{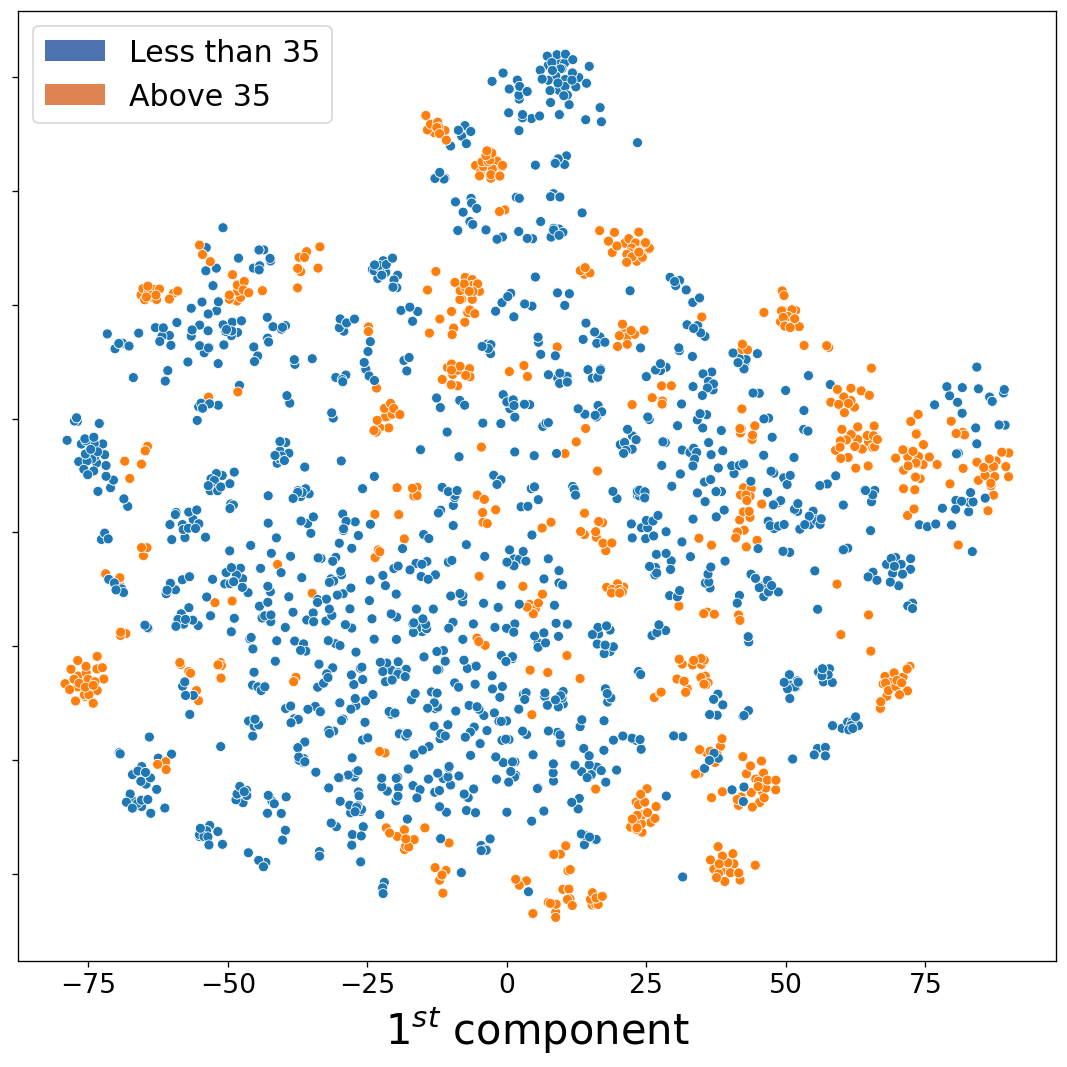} \\
    \scriptsize (c) ECG + clinic. letters\\ \scriptsize without augmentations
  \end{tabular}
  \begin{tabular}[b]{c}
    \includegraphics[width=.22\linewidth]{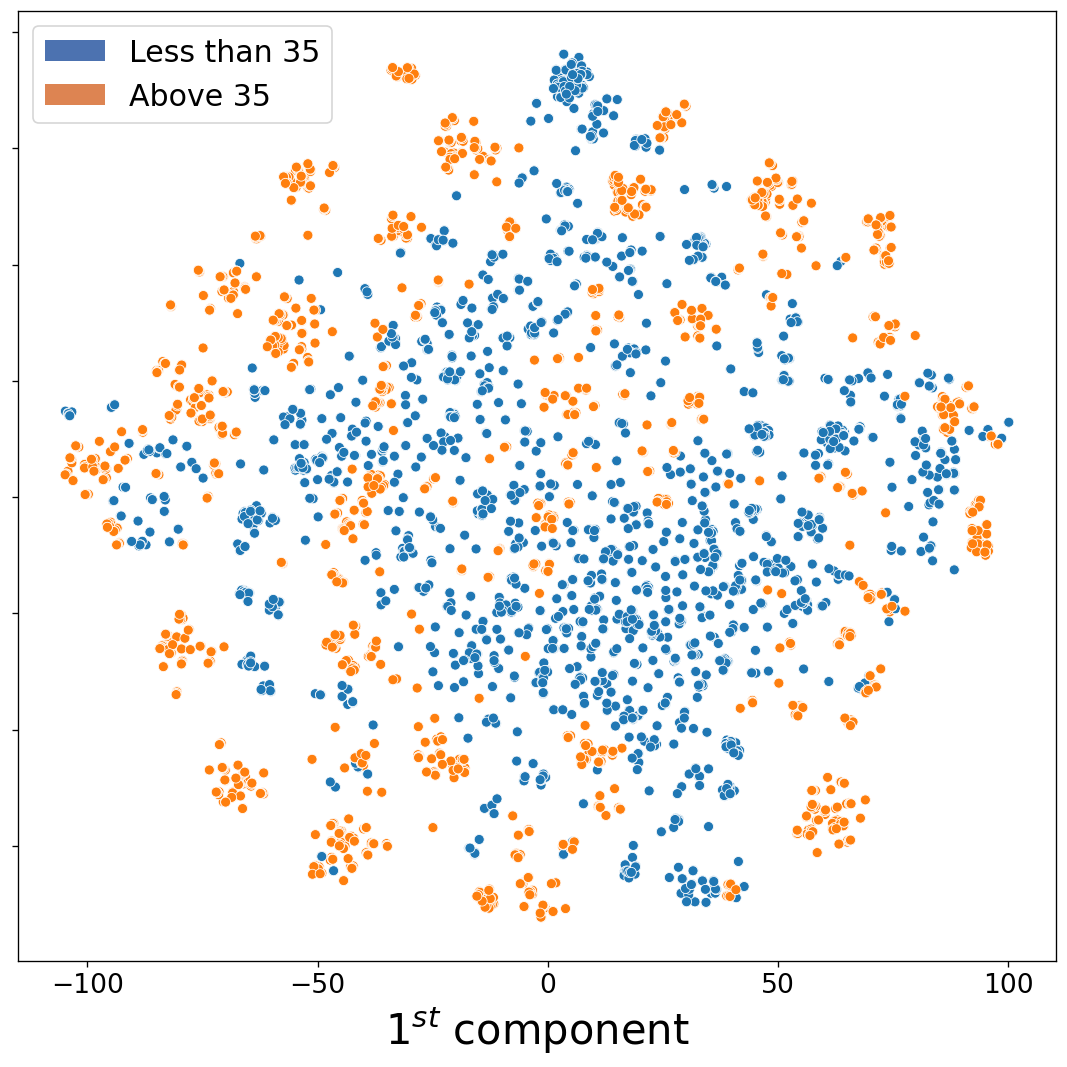} \\
    \scriptsize (d) ECG + clinic. letters\\ \scriptsize without augmentations
  \end{tabular}
  \caption{Comparison of t-SNE visualizations on the tangent space}
  \label{fig:TSNE}
\end{figure}

For the augmentation of clinical data, a similar weighted mean approach in Equation \ref{eq:meanRiemannian} is applied to the vectors that contain clinical letter information, without any distance function. The two vectors were averaged according to the specified weight $\alpha$. All the augmentation is applied in the same method in order to maintain consistency in the value of $\alpha$ for both ECG and clinical data augmentations of the same data. To qualitatively validate the effectiveness of this approach, we employed $t\text{-}SNE$ visualizations \cite{van2008visualizing} on the tangent space, comparing representations using only original data and those combined with the mixed data, as illustrated in Figure \ref{fig:TSNE}.

\subsection{Prediction Models}
\subsection*{Regression Models}

To compare the effect of fusing different input types, we employed two regression models: Support Vector Machine (SVM) and Logistic Regression (LR). Ablation studies were conducted using the SVM model, with the LR model serving as the baseline. The data was split into training and testing sets 100 times using a pseudo-randomized, stratified patient leave-out evaluation. This method ensured that the testing set was representative of all classes by randomly selecting one patient from each class for the testing set, while the remaining patients populated the training set. For each run, the SVM model was trained on the training set and then tested on the corresponding testing set, which did not include any data from the patients in the training set. This stratified patient leave-out method enhances the model's ability to predict unseen data by reducing bias \cite{cawley2010over}. The same patient split strategy was used to ensure consistency across different results. 

\begin{figure}[!htp]
  \centering
  \begin{tabular}[b]{c}
    \includegraphics[height=.3\linewidth]{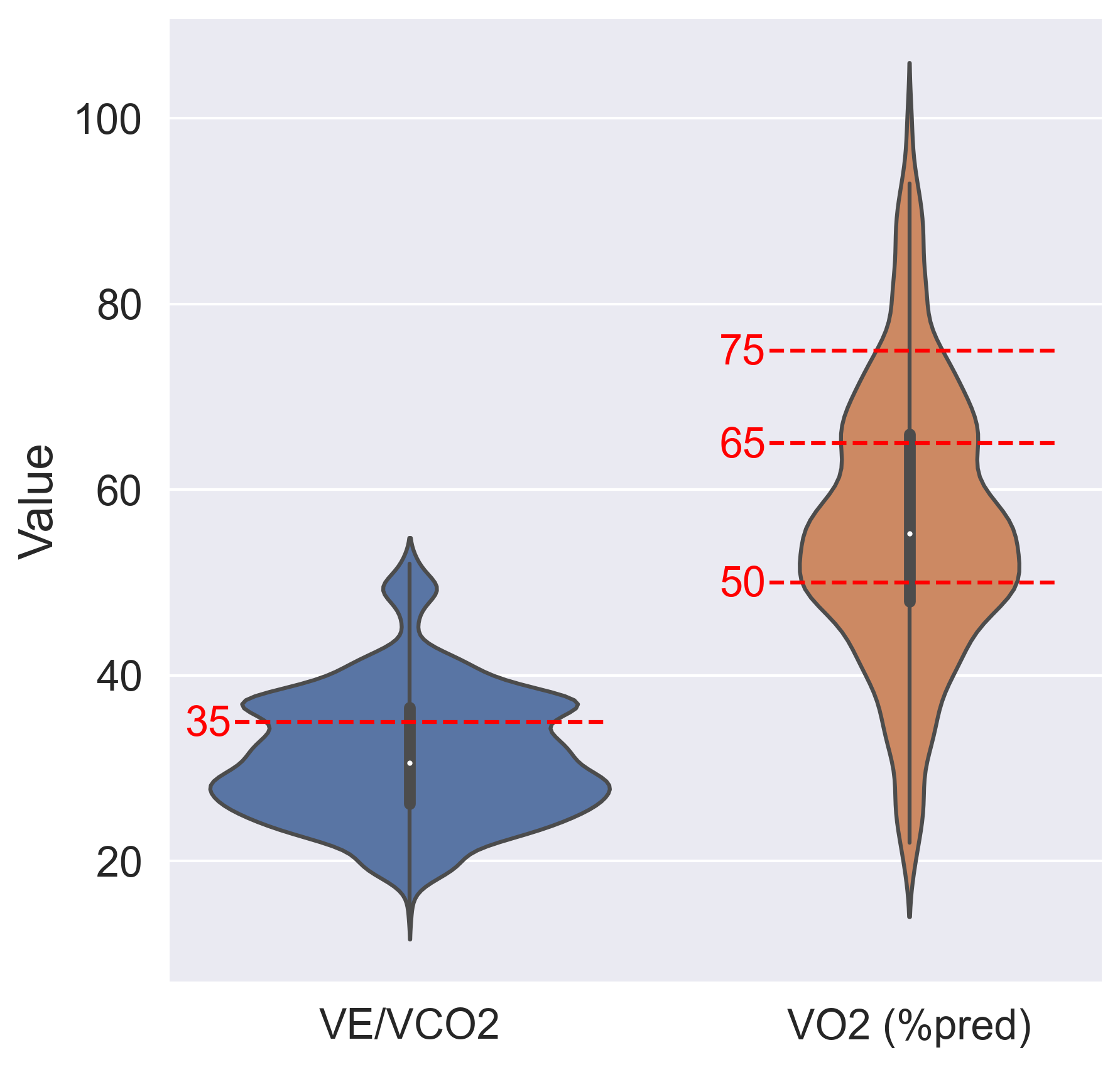} \\
  \end{tabular}
  \begin{tabular}[b]{c}
    \includegraphics[height=.3\linewidth]{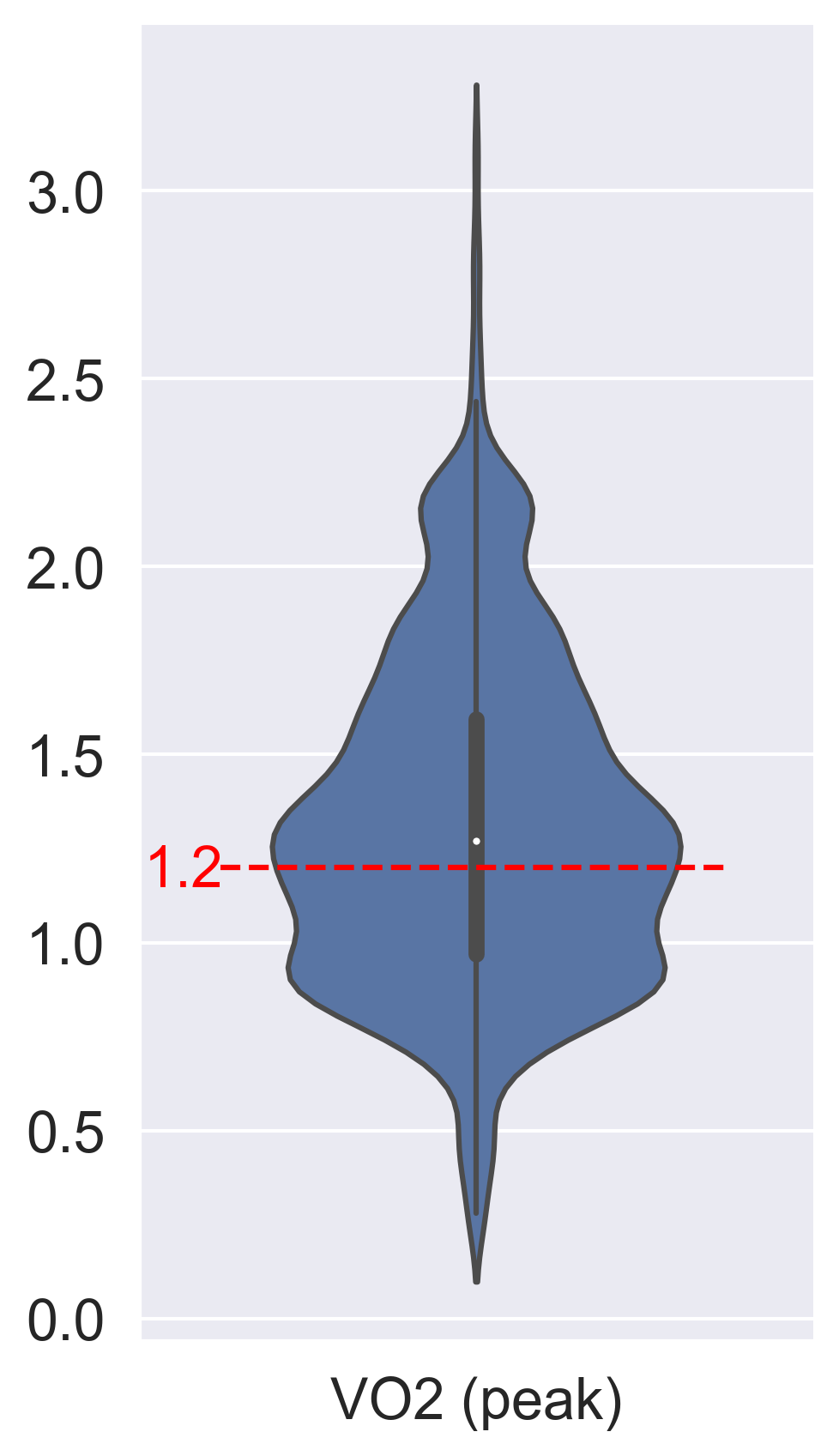} \\
  \end{tabular}
  \caption{Data distribution of each class, with CPET groups}
  \label{fig:his}
\end{figure}

\subsection*{Classification Model}

The regression problem was transformed into a classification problem by assigning clinical groups for each label of $VE/VCO_2$, $VO_2$ (\%pred) and $VO_2$ (peak). Specifically, $VE/VCO_2$ and $VO_2$ (peak) variables were categorized into two groups, while $VO_2$ (\%pred) variable was divided into four groups. As demonstrated in Figure \ref{fig:his}, the separation of the groups is indicated by dashed red lines. A series of ablation studies were conducted employing an SVM model to examine the impact of various factors, including the fusion of different input types and the application of different augmentation techniques. To ensure the reliability and reproducibility of the results, the same patient split strategy and training setup were consistently used across all experiments. The performance of the classification model was evaluated using multiple metrics, including accuracy, Area Under the Curve (AUC), and F1 macro score. These metrics were reported alongside their corresponding mean and standard deviation values to provide a comprehensive assessment.

\section{Results}
\subsection{Regression}

To establish a baseline, we first trained logistic regression models using standard ECG measurements from vendor reports (Marquette 12SL ECG \cite{healthcare2008marquette}), including PR intervals, QRS durations, and ventricular rates. Using the vendor manual \cite{healthcare2008marquette}, similar calculations performed on the extracted ECG signals to derive calculated ECG features. We evaluated the model performance using multiple metrics that involve the coefficient of determination (R²), adjusted R², root mean square error (RMSE), and correlation coefficients (r). In Equation \ref{eq:r2}, $\hat{y}$ represents the predicted y value and $\bar{y}$ represents the mean of the y values.

\begin{equation}
\label{eq:r2}
R^2=1-\frac{\text{sum squared regression (SSR)}}{\text{total sum of squares (SST)}} =1-\frac{\sum({y_i}-\hat{y_i})^2}{\sum(y_i-\bar{y})^2}
\end{equation}

Table \ref{tab:regressionTable} summaries all the evaluation metrics for each prediction label, $VE/VCO_2$, $VO_2$ (\%pred) and $VO_2$ (peak). Kernel Density Estimations (KDEs) are also employed to illustrate the distribution of both the actual and predicted values. For the baseline model, it was evident that the predicted values were constrained to a relatively limited range, whereas the actual values demonstrated a substantially broader distribution.
The results indicate that there was no significant correlation between the model's predictions, based on derived ECG parameters, and the actual exercise values. Although these ECG parameters are frequently reported in ECG PDF documents, they do not provide enough information to effectively train a regression model to predict CPET measurements.\\

\begin{table}[h]
\centering
\caption{\centering Predicting Cardiopulmonary Exercise Biomarkers using our proposed approach and SVM}
\resizebox{\textwidth}{!}{%
\begin{tabular}{@{}c|cccc|c|c|c@{}}
\hline
\rowcolor[HTML]{FFFFFF} 
 & \textbf{$R^2$ $\uparrow$} & \textbf{Adjusted $R^2$} & \textbf{RMSE $\downarrow$} & \textbf{r $\uparrow$} & \textbf{Predicted Label} & \textbf{Model} & \textbf{Input Data} \\ \hline
\rowcolor[HTML]{CCCCCC} 
\cellcolor[HTML]{FFFFFF}Vendor ECG features* & -0.080 & -0.086 & 8.695 & 0.196 & \cellcolor[HTML]{FFDAC0}$VE/VCO_2$ & LR & PR interval, QRS duration, Vent. rate \\
\cellcolor[HTML]{FFFFFF}Calculated ECG features & -0.171 & -0.178 & 9.056 & -0.025 & \cellcolor[HTML]{FFDAC0}$VE/VCO_2$ & LR & PR interval, QRS duration, Vent. rate \\
\rowcolor[HTML]{CCCCCC} 
\cellcolor[HTML]{FFFFFF}Vendor ECG features & -0.109 & -0.118 & 10.613 & 0.247 & \cellcolor[HTML]{FFDAC0}$VE/VCO_2$ & SVM & PR interval, QRS duration, Vent. rate \\
\multicolumn{1}{c|}{\cellcolor[HTML]{FFFFFF}Tangent Space*} & \multicolumn{1}{c}{0.011} & 0.082 & 8.178 & 0.349 & \cellcolor[HTML]{FFDAC0}$VE/VCO_2$ & SVM & ECGs \\
\rowcolor[HTML]{CCCCCC} 
\multicolumn{1}{c|}{\cellcolor[HTML]{FFFFFF}Tangent Space} & \multicolumn{1}{c}{\underline{0.111}} & -0.084 & \underline{7.754} & \underline{0.478} & \cellcolor[HTML]{FFDAC0}$VE/VCO_2$ & SVM & ECGs + clinical letters \\
\multicolumn{1}{c|}{\cellcolor[HTML]{FFFFFF}Tangent Space (aug)*} & \multicolumn{1}{c}{\textbf{0.130}} & -0.060 & \textbf{7.670} & \textbf{0.491} & \cellcolor[HTML]{FFDAC0}$VE/VCO_2$ & SVM & ECGs + clinical letters \\
\rowcolor[HTML]{CCCCCC} 
\cellcolor[HTML]{FFFFFF}Vendor ECG features* & -0.093 & -0.099 & 9.221 & -0.001 & \cellcolor[HTML]{BBB4D9}$VO_2$ (\%pred) & LR & PR interval, QRS duration, Vent. rate \\
\cellcolor[HTML]{FFFFFF}Calculated ECG features & -0.106 & -0.112 & 9.274 & -0.021 & \cellcolor[HTML]{BBB4D9}$VO_2$ (\%pred) & LR & PR interval, QRS duration, Vent. rate \\
\rowcolor[HTML]{CCCCCC} 
\cellcolor[HTML]{FFFFFF}Vendor ECG features & -0.177 & -0.183 & 9.567 & 0.188 & \cellcolor[HTML]{BBB4D9}$VO_2$ (\%pred) & SVM & PR interval, QRS duration, Vent. rate \\
\multicolumn{1}{c|}{\cellcolor[HTML]{FFFFFF}Tangent Space*} & \multicolumn{1}{c}{0.218} & 0.046 & 7.892 & 0.550 & \cellcolor[HTML]{BBB4D9}$VO_2$ (\%pred) & SVM & ECGs \\
\rowcolor[HTML]{CCCCCC} 
\multicolumn{1}{c|}{\cellcolor[HTML]{FFFFFF}Tangent Space} & \multicolumn{1}{c}{\textbf{0.263}} & 0.101 & \textbf{7.662} & \underline{0.658} & \cellcolor[HTML]{BBB4D9}$VO_2$ (\%pred) & SVM & ECGs + clinical letters \\
\multicolumn{1}{c|}{\cellcolor[HTML]{FFFFFF}Tangent Space (aug)*} & \multicolumn{1}{c}{\underline{0.243}} & 0.076 & \underline{7.769} & \textbf{0.662} & \cellcolor[HTML]{BBB4D9}$VO_2$ (\%pred) & SVM & ECGs + clinical letters \\
\rowcolor[HTML]{CCCCCC} 
\cellcolor[HTML]{FFFFFF}Vendor ECG features* & -0.054 & -0.060 & 0.444 & 0.051 & \cellcolor[HTML]{F8CFD0}$VO_2$ (peak) & LR & PR interval, QRS duration, Vent. rate \\
\cellcolor[HTML]{FFFFFF}Calculated ECG features & -0.058 & -0.064 & 0.445 & 0.030 & \cellcolor[HTML]{F8CFD0}$VO_2$ (peak) & LR & PR interval, QRS duration, Vent. rate \\
\rowcolor[HTML]{CCCCCC} 
\cellcolor[HTML]{FFFFFF}Vendor ECG features & 0.024 & 0.018 & 0.427 & 0.243 & \cellcolor[HTML]{F8CFD0}$VO_2$ (peak) & SVM & PR interval, QRS duration, Vent. rate \\
\multicolumn{1}{c|}{\cellcolor[HTML]{FFFFFF}Tangent Space*} & 0.303 & 0.149 & 0.358 & 0.573 & \cellcolor[HTML]{F8CFD0}$VO_2$ (peak) & SVM & ECGs \\
\rowcolor[HTML]{CCCCCC} 
\cellcolor[HTML]{FFFFFF}Tangent Space & \textbf{0.395} & 0.261 & \textbf{0.333} & \underline{0.660} & \cellcolor[HTML]{F8CFD0}$VO_2$ (peak) & SVM & ECGs + clinical letters \\
\multicolumn{1}{c|}{\cellcolor[HTML]{FFFFFF}Tangent Space (aug)*} & \multicolumn{1}{c}{\textbf{0.395}} & 0.261 & \textbf{0.333} & \textbf{0.666} & \cellcolor[HTML]{F8CFD0}$VO_2$ (peak) & SVM & ECGs + clinical letters \\ \hline
\end{tabular}%
}
\label{tab:regressionTable}
\end{table}

\begin{figure}[!h]
  \centering
  \begin{tabular}[b]{c}
    \includegraphics[width=.29\linewidth]{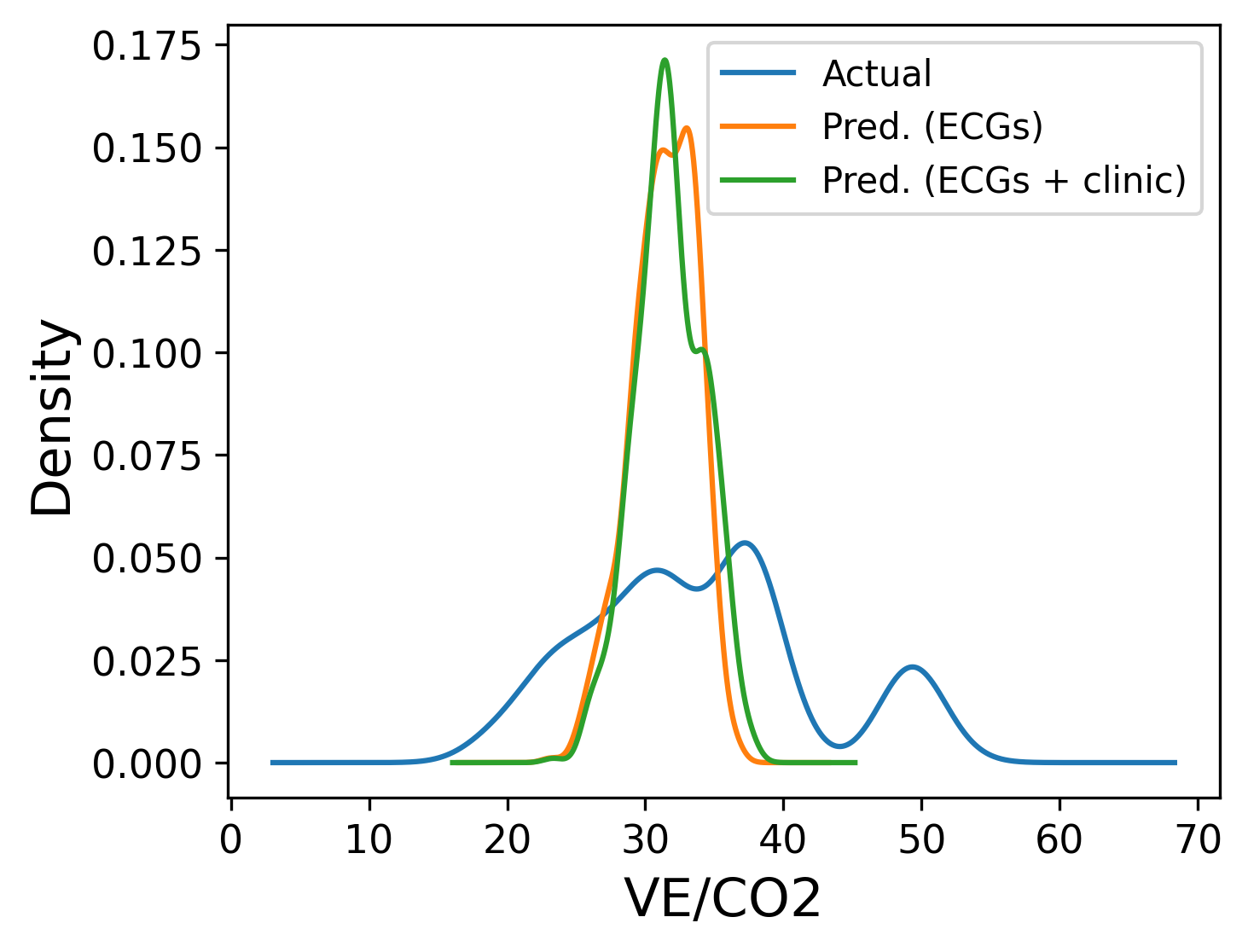} \\
    \small (a) $VE/VCO_2$
  \end{tabular}
  \begin{tabular}[b]{c}
    \includegraphics[width=.29\linewidth]{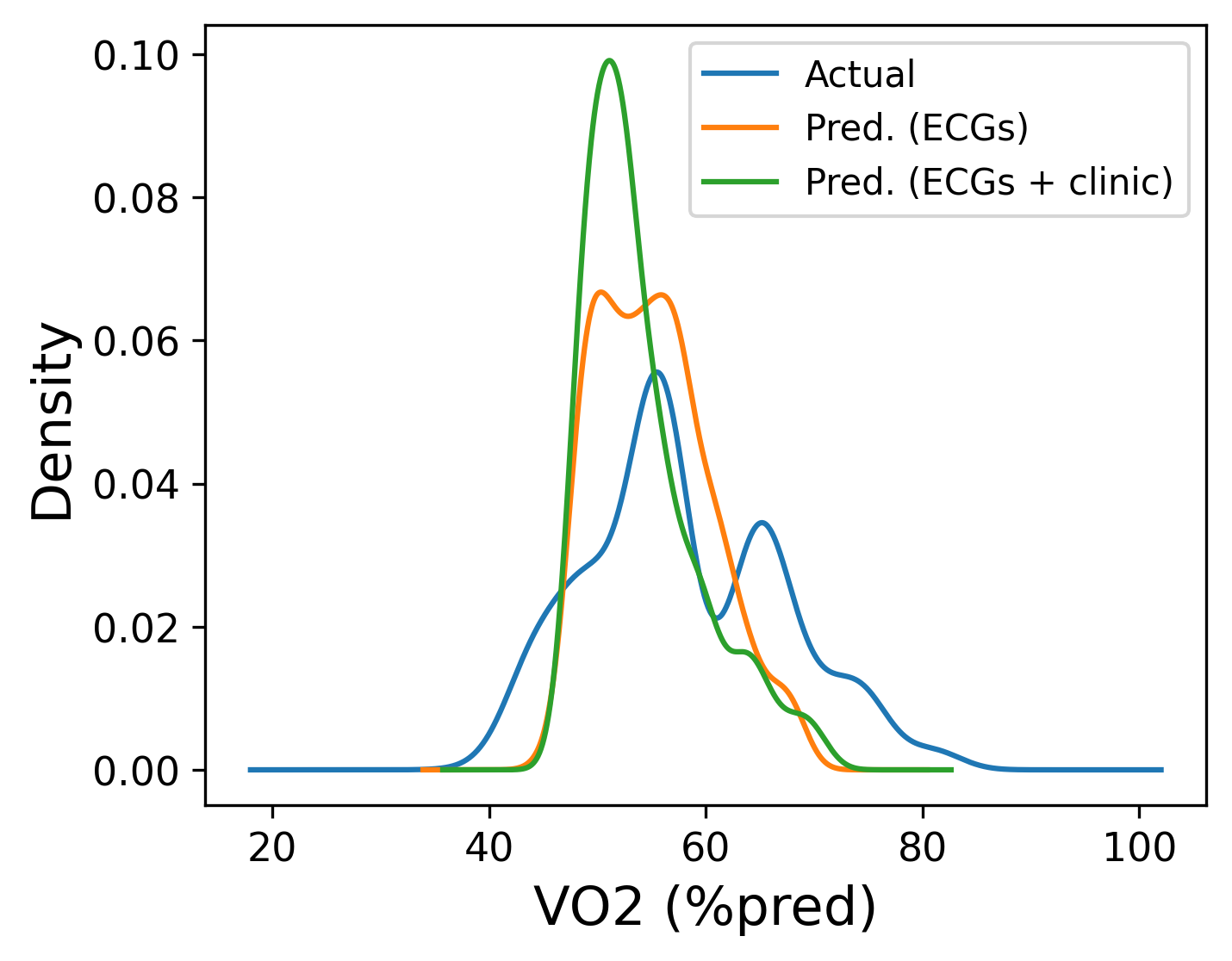} \\
    \small (b) $VO_2$ (\%pred)
  \end{tabular}
  \begin{tabular}[b]{c}
    \includegraphics[width=.29\textwidth]{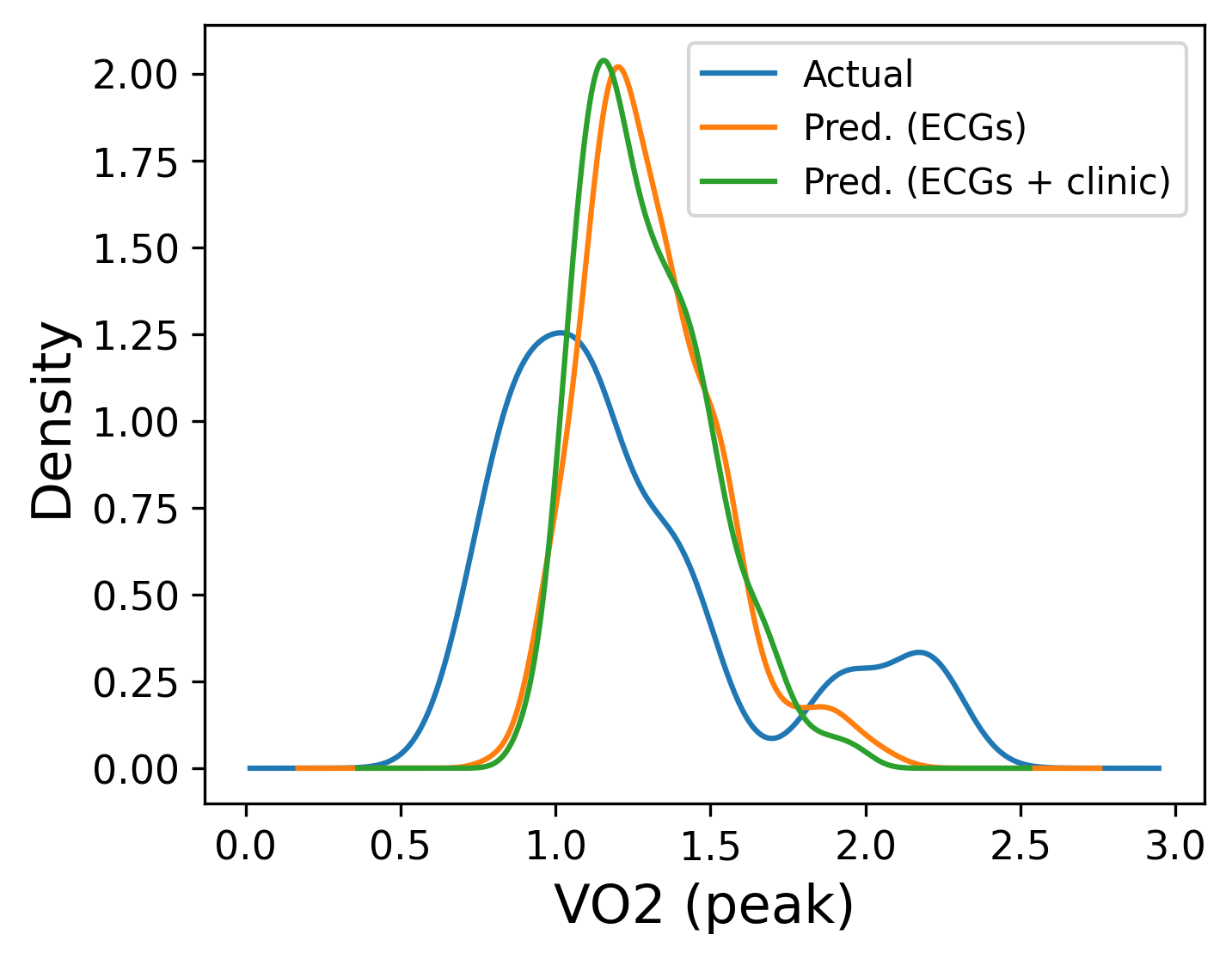} \\
    \small (b) $VO_2$ (peak)
  \end{tabular}
  \caption{Kernel Density Estimations (KDEs) of the SVM model}
  \label{fig:kdeSVM}
\end{figure}

As there was no correlation between the prediction and the actual value, and all predictions fall within a narrow range, we explored tangent space mappings. We first calculated covariance matrices of the ECG signals and then map all to the tangent space using Riemannian distance. It yielded better results and also a better density plot of the predictions, as can be seen in Table \ref{tab:regressionTable}. Table \ref{tab:regressionTable} summaries all the evaluation metrics for each prediction label, $VE/VCO_2$, $VO_2$ (\%pred) and $VO_2$ (peak). The correlation between the model's predictions and the actual exercise values improved significantly, reaching 0.57 for $VO_2$ (peak). This performance was notably better than using parameters commonly reported in ECG PDF documents, such as PR interval and QRS duration. KDEs plotted in Figure \ref{fig:kdeSVM} show the distribution of actual and predicted values. These plots indicate  that all predictions are much closer to the distribution of actual exercise values, demonstrating the effectiveness of our approach.

Table \ref{tab:regressionTable} also summarizes the results obtained by combining  information from clinical letters with ECG signals projected onto tangent space. All patient history information regarding interventions, diagnoses, and medications was utilized. Specifically, the frequency count of each unique word was computed and transformed into a numerical matrix of word counts. These matrices are then concatenated with the tangent space mapping matrices to feed the model. This data fusion approach achieved better results in terms of all the evaluation metrics. The correlation between the model's prediction and the actual exercise value significantly improved, reaching 0.66 for $VO_2$ (peak). KDEs plotted in Figure \ref{fig:kdeSVM} illustrate the distribution of the actual and predicted values, showing that the predictions are much closer to the distribution of actual exercise values. These plots also demonstrated a positive alignment with the improved results. Additionally, the covariance augmentation technique was also applied on the fusion of both ECG and clinical letters in order to further enhance the results.\\ 

\begin{figure}[!h]
  \centering
  \begin{tabular}[b]{c}
    \includegraphics[width=.3\linewidth]{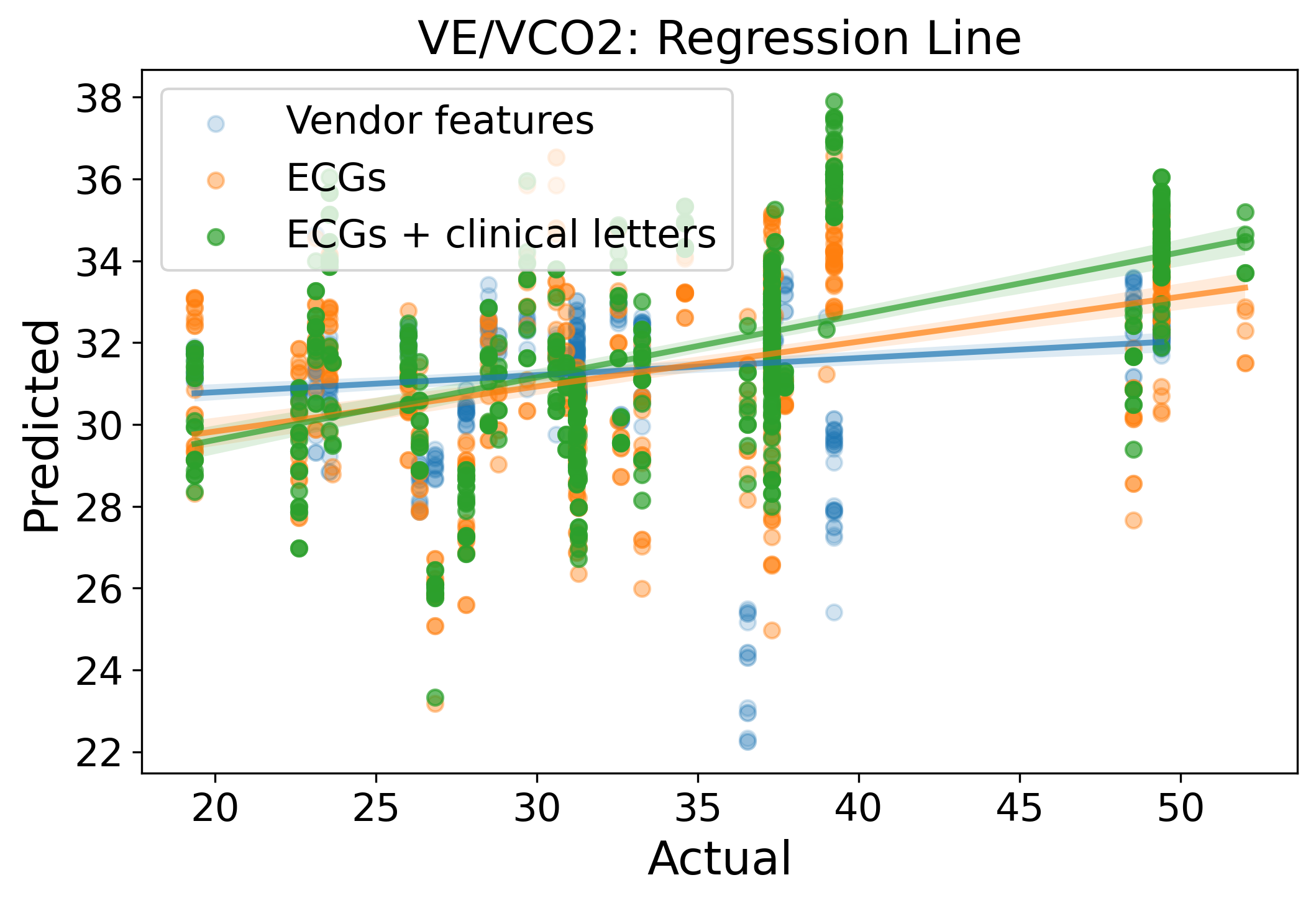} \\
    \small (a) $VE/VCO_2$
  \end{tabular}
  \begin{tabular}[b]{c}
    \includegraphics[width=.3\linewidth]{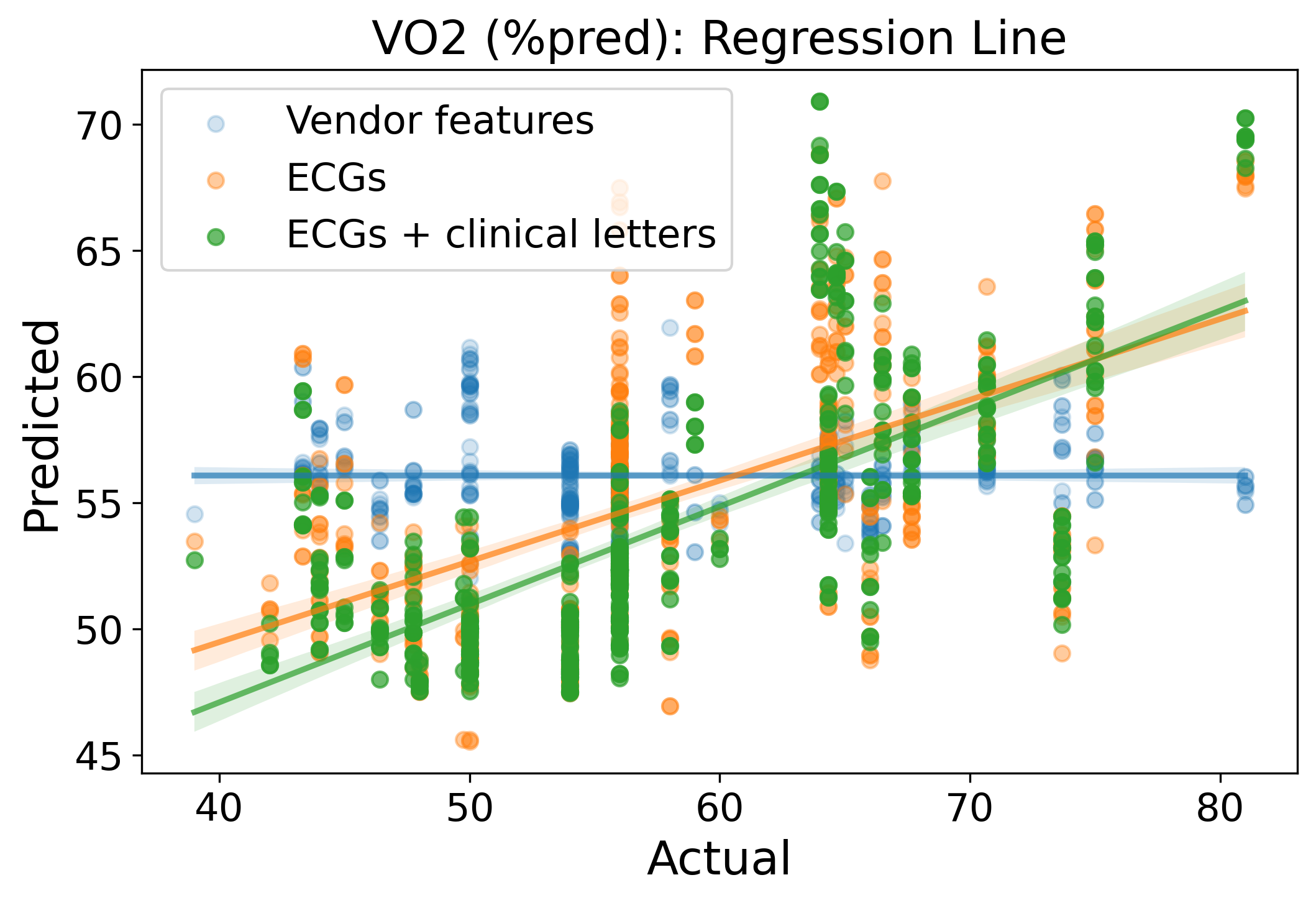} \\
    \small (b) $VO_2$ (\%pred)
  \end{tabular}
  \begin{tabular}[b]{c}
    \includegraphics[width=.3\textwidth]{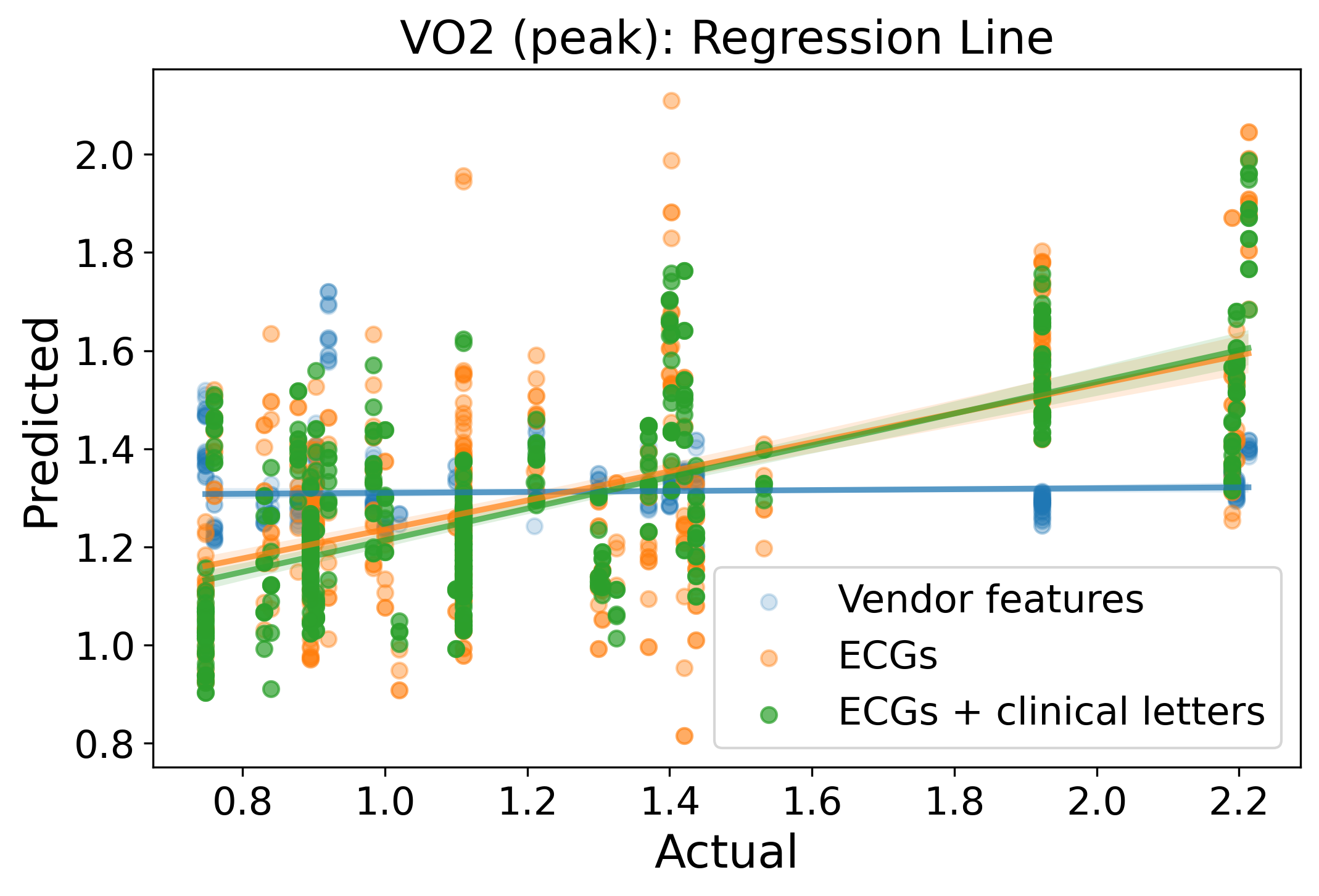} \\
    \small (b) $VO_2$ (peak)
  \end{tabular}
  \caption{\centering Predicting cardiopulmonary exercise biomarkers: comparison of predicted vs. real values with fitted regression lines}
  \label{fig:regLine}
\end{figure}

In order to provide a more concise summary of the each step, a regression plot for each label was also provided in Figure \ref{fig:regLine}. The plots (denoted with * in Table \ref{tab:regressionTable}) illustrate three regression lines for each step, starting from the vendor features and proceeding to the combination of ECGs and clinical letters. These plots illustrate the capability of the model and provide a comprehensive overview of the improvements achieved.

\vspace{-7mm}

\subsection{Classification}

In order to transform the regression problem into a relatively simple classification problem, clinical groups were determined for each label of $VE/VCO_2$, $VO_2$ (\%pred) and $VO_2$ (peak) with data distribution of each class as shown in Figure \ref{fig:his}. Initial experiments were carried out without any augmentation in order to establish a baseline. The baseline SVM model was employed with two separate inputs: ECGs alone, and ECGs in conjunction with clinical letters. Using tangent mappings derived from the ECGs as inputs to the SVM model did not yield the best results. However, a notable improvement (up to \%8 on accuracy and AUC, \%5 on F1 score) was observed when the ECG data combined with the clinical letter data, which proved to be advantageous in terms of the classification metrics. In order to enhance the result and the balance of the class distribution, the experiments were repeated with the augmentations, both on ECGs and clinical letters.\\

\definecolor{Negroni}{rgb}{1,0.854,0.752}
\definecolor{Silver}{rgb}{0.8,0.8,0.8}
\definecolor{LavenderGray}{rgb}{0.733,0.705,0.85}
\definecolor{Azalea}{rgb}{0.972,0.811,0.815}
\begin{table}[h]
\centering
\caption{\centering Classification results using the grouping labels shown at figure \ref{fig:his} and a SVM model}
\resizebox{1\textwidth}{!}{
\begin{tblr}{
  cells = {c},
  row{3} = {Silver},
  row{5} = {Silver},
  row{7} = {Silver},
  row{9} = {Silver},
  row{11} = {Silver},
  row{13} = {Silver},
  row{15} = {Silver},
  row{17} = {Silver},
  row{19} = {Silver},
  cell{1}{1} = {r=2}{},
  cell{1}{2} = {c=2}{0.187\linewidth},
  cell{1}{4} = {c=2}{0.232\linewidth},
  cell{1}{6} = {r=2}{},
  cell{1}{7} = {r=2}{},
  cell{1}{8} = {r=2}{},
  cell{3}{1} = {Negroni},
  cell{4}{1} = {Negroni},
  cell{5}{1} = {Negroni},
  cell{6}{1} = {Negroni},
  cell{7}{1} = {Negroni},
  cell{8}{1} = {LavenderGray},
  cell{9}{1} = {LavenderGray},
  cell{10}{1} = {LavenderGray},
  cell{11}{1} = {LavenderGray},
  cell{12}{1} = {LavenderGray},
  cell{13}{1} = {Azalea},
  cell{14}{1} = {Azalea},
  cell{15}{1} = {Azalea},
  cell{16}{1} = {Azalea},
  cell{17}{1} = {Azalea},
  vlines,
  hline{1,3,8,13,18} = {-}{},
  hline{2} = {2-5}{},
}
{\textbf{Predicted}\\\textbf{Label}} & \textbf{Input} &  & \textbf{Augmentations} &  & \textbf{Accuracy} & \textbf{AUC} & \textbf{F1 macro}\\
 & ECGs & Clinical letters & ECGs & Clinical letters &  &  & \\
$VE/VCO_2$ & ✓ &  &  &  & 0.672 ± 0.212 & 0.577 ± 0.141 & 0.534 ± 0.181\\
$VE/VCO_2$ & ✓ &  & covariance &  & 0.704 ± 0.195 & 0.622 ± 0.183 & 0.572 ± 0.197\\
$VE/VCO_2$ & ✓ & ✓ &  &  & 0.696 ± 0.227 & 0.655 ± 0.179 & 0.587 ± 0.236\\
$VE/VCO_2$ & ✓ & ✓ & covariance & simple & 0.675 ± 0.240 & 0.620 ± 0.179 & 0.541 ± 0.238\\
$VE/VCO_2$ & ✓ & ✓ & covariance & covariance & \textbf{0.738 ± 0.192} & \textbf{0.663 ± 0.180} & \textbf{0.617 ± 0.206}\\
$VO_2$ (\%pred) & ✓ &  &  &  & 0.302 ± 0.163 & 0.506 ± 0.225 & 0.229 ± 0.127\\
$VO_2$ (\%pred) & ✓ &  & covariance &  & 0.324 ± 0.132 & 0.520 ± 0.225 & 0.236 ± 0.106\\
$VO_2$ (\%pred) & ✓ & ✓ &  &  & 0.316 ± 0.124 & 0.529 ± 0.229 & 0.239 ± 0.093\\
$VO_2$ (\%pred) & ✓ & ✓ & covariance & simple & 0.312 ± 0.116 & 0.537 ± 0.255 & 0.239 ± 0.095\\
$VO_2$ (\%pred) & ✓ & ✓ & covariance & covariance & \textbf{\textbf{0.331 ± 0.124}} & \textbf{\textbf{0.540 ± 0.245}} & \textbf{\textbf{0.257 ± 0.093}}\\
$VO_2$ (peak) & ✓ &  &  &  & 0.614 ± 0.137 & 0.585 ± 0.162 & 0.535 ± 0.167\\
$VO_2$ (peak) & ✓ &  & covariance &  & 0.657 ± 0.150 & 0.620 ± 0.126 & 0.553 ± 0.152\\
$VO_2$ (peak) & ✓ & ✓ &  &  & 0.693 ± 0.163 & 0.601 ± 0.136 & 0.525 ± 0.169\\
$VO_2$ (peak) & ✓ & ✓ & covariance & simple & 0.642 ± 0.111 & 0.607 ± 0.093 & 0.544 ± 0.120\\
$VO_2$ (peak) & ✓ & ✓ & covariance & covariance & \textbf{0.708 ± 0.091} & \textbf{0.658 ± 0.134} & \textbf{0.617 ± 0.142}
\end{tblr}
}
\label{tab:regToCSVM}
\end{table}

A comprehensive summary of all results is provided in Table \ref{tab:regToCSVM}, with the most optimal results indicated in bold. It was observed that applying covariance augmentations only to ECGs was able to increase accuracy, AUC and the F1 score for all classes by up to \%4. When clinical letters were incorporated, a similar outcome was observed with the covariance augmentations applied both to ECGs and clinical letters. It was able to increase accuracy up to \%4, AUC up to \%5 and the F1 score up to \%9 for all classes. However, using simple augmentation techniques as opposed to the covariance augmentations on clinical letters resulted in poor performances, as presented in Table \ref{tab:regToCSVM}. It also demonstrates the importance of the augmentation technique that is tailored to the particular problem.\\

Our approach to fuse information derived from ECGs and clinical letters, and to use a sophisticated augmentation technique yielded the best results. The utilisation of Riemannian geometry augmentations on the covariance matrices of ECGs and clinical letters has been shown to produce features that are more coherent in comparison to those obtained through simple augmentation techniques. The efficacy of our methodology was evaluated with ablation studies, which demonstrated that the integration of ECGs and clinical data yielded the best results. 

\section{Discussion}

Medical data analysis, particularly concerning CHD, poses challenges due to the inherent complexity and imbalanced nature of the data. Large machine learning models often struggle in such contexts, primarily due to their sensitivity to data distribution and inefficiency in learning from datasets with limited examples. Our approach leverages the geometric properties of Riemannian spaces, offering a more robust and discriminative feature space for machine learning models. By utilizing the non-Euclidean nature of the data, our method captures intrinsic geometrical structures that are frequently overlooked by conventional methods. Our study demonstrated promising results in predicting surrogate mortality for patients with congenital heart disease (CHD). The proposed projection of augmented covariance matrices to Riemannian spaces significantly improved performance with small and extremely imbalanced 12-lead ECG data. The experimental results support the hypothesis that the proposed solution is effective for both regression and classification problems.\\

The integration of clinical letters with ECGs and the utilization of Riemannian geometry augmentations yielded the best results. Clinical letters often contain vital information about a patient's medical history, symptoms, and other relevant details not captured by ECG signals alone. Integrating this textual information with ECG data provides a more comprehensive and informative feature set. The experimental results demonstrate the efficiency of the proposed augmentation technique in generating more coherent features.\\

The potential of developing clinical models for predicting mortality and disease complexity in CHD patients has been previously explored in the literature \cite{diller2019machine, mayourian2024electrocardiogram}. Previous research has explored clinical models for predicting mortality and disease complexity in CHD patients. Diller et al. \cite{diller2019machine} focused on categorizing diagnosis and disease complexity using data from over 10,000 patients, while Mayourian et al. \cite{mayourian2024electrocardiogram} achieved an AUC of approximately 79\% in predicting mortality based on more than 39,000 patients and 100,000 ECGs. These studies highlight the difficulty in stratifying risk for CHD patients due to the low prevalence of adverse events.\\

To address this challenge, we innovatively adopted CPET outcomes to provide a more dynamic evaluation of patient condition and risk. To our knowledge, this represents the first attempt to build a sophisticated machine learning model predicting CPET results as outcome variables. The improved correlation between our model's predictions and actual exercise values, particularly for $VO_2$ (peak), demonstrates the potential of our approach in clinical settings even with limited data.\\

Future research directions include exploring advanced natural language processing techniques to extract richer information from clinical letters and investigating the integration of multi-modal data sources, such as imaging, to provide more comprehensive patient profiles.\\

In conclusion, our study demonstrates the significant potential of combining clinical letters with ECG data and leveraging Riemannian geometry to enhance predictive performance in CHD patients. The use of CPET as an outcome variable provides a dynamic and physiologically relevant endpoint that better reflects functional capacity and cardiovascular reserve compared to static measurements alone. This approach effectively addresses challenges posed by small, imbalanced datasets while providing more accurate patient risk assessments.

\section*{Acknowledgements}

Fani Deligianni is supported by funding from EPSRC (EP/W01212X/1) and Academy of Medical
Sciences (NGR1/1678). She is also a member of the research team for NIHR (NIHR158303).

\section*{Patient Consent Statement}

Study approval was obtained from the Institutional Governance Division of the NHS Golden Jubilee National Hospital and the Health Research Authority (23/SC/0436, IRAS project ID: 335717).

\printbibliography
\end{document}